\pdfoutput=1
\documentclass{jku-icg-thesis}

\usepackage{booktabs}
\usepackage{csquotes}
\usepackage{graphicx}
\usepackage{subcaption}
\usepackage{xcolor}
\usepackage{tikz}
\usetikzlibrary{positioning, arrows.meta, arrows, backgrounds, fit}

\usepackage[
    style=numeric,
    backend=biber, 
    ]{biblatex}

\usepackage{kantlipsum}

\usepackage{listings}
\lstdefinestyle{promptfile}{
    basicstyle=\ttfamily\footnotesize,
    breaklines=true,
    breakatwhitespace=false,
    columns=fullflexible,
    keepspaces=true,
    frame=single,
    framesep=6pt,
    rulecolor=\color{black!30},
    xleftmargin=8pt,
    xrightmargin=0pt,
    aboveskip=1em,
    belowskip=1em,
    literate={ä}{{\"a}}1 {ö}{{\"o}}1 {ü}{{\"u}}1 {ß}{{\ss}}1
}

\newcommand{\Autoref}[1]{%
    \begingroup
    \def\chapterautorefname{Chapter}%
    \def\sectionautorefname{Section}%
    \def\subsectionautorefname{Subsection}%
    \def\subsubsectionautorefname{Subsection}%
    \def\figureautorefname{Figure}%
    \def\tableautorefname{Table}%
    \def\equationautorefname{Equation}%
    \def\appendixautorefname{Appendix}%
    \autoref{#1}%
    \endgroup
}

\thesissetup{
    subject = ai,       
    study = msc,        
    location = Linz,
    month = 10,         
    year = 2026,
}

\input glyphtounicode
\begin{document}

\begin{titlepage}
    \centering
    \vspace*{2cm}
    {\fontsize{22}{28}\selectfont\bfseries Predicting Void Points in Self-Occluding Forest Volumes\par}
    \vspace{0.8em}
    {\large Improving the DeepForest approach with Void Point Volumes\par}
    \vspace{2em}
    \begin{tabular}{ccc}
        \large Benjamin Bachmayr & \large Mohamed Youssef & \large Oliver Bimber
    \end{tabular}\par
    \vspace{1em}
    Institute of Computer Graphics\\
    Johannes Kepler University Linz, Austria\par
    \vspace{1em}
    October 2026\par
\end{titlepage}

\pagenumbering{roman}

\pdfbookmark[1]{Abstract}{sec:abstract-en}
\chapter*{Abstract}

Synthetic Aperture Imaging (SAI) lets a camera drone see into a forest canopy by warping many displaced views onto a common depth and averaging them. Occlusion is suppressed this way, but it is not removed. A position deep inside a canopy that only few camera rays ever reached still receives a value during reconstruction, and that value looks like real matter to any filter working on intensity alone. Aggregating occupied voxels over a stand therefore overestimates biomass in one direction only. The learning-based pipeline of the Deep Forest project cannot tell the two apart, because its training pairs are anchored on points that carry a measured reflectance. Empty air produces no such anchor and is dropped from the training set, so the network is never asked the question it is later expected to answer.

This thesis changes the labels rather than the imaging. Each simulated scene is rendered twice: once as the $9\times9$ grid of top-down captures that becomes the focal stack the network sees, and once as 550 depth-clipped slabs that read occupancy straight off the scene geometry. Void space becomes an explicit negative class instead of a gap. A Gazebo pipeline generates procedural stands of 80 to 200 trees across three species, a pre-trained Pix2Pix generator translates the rendered RGB into a synthetic near-infrared band, and all three channels are pushed through Light Field Rendering into aligned focal stacks of 550 planes.

Four architectures were compared on a held-out scene, and the ordering does not follow the architectural progression. The plain 2D U-Net scores highest at an IoU of 0.6290, ahead of the volumetric 3D U-Net at 0.5707. Against established reconstruction methods the gap is wide. COLMAP peaks at 0.3446 and DepthAnything3 at 0.2959, each after its own best dilation sweep, while an RGB-input variant of the volumetric architecture reaches 0.6514 with neither dilation nor any per-scene calibration. Both baselines fail for structural reasons rather than through poor tuning: they reconstruct visible surfaces, and the interior of a closed canopy offers none.

No per-voxel reference exists for any real capture, so the transfer to real drone data is described qualitatively and its accuracy remains unmeasured.

\ClearPage

\pdfbookmark[1]{Contents}{sec:toc}
\graphicspath{{figures/}{resources/images/}}
\ClearPage

\setcounter{page}{1}
\pagenumbering{arabic}


\chapter{Introduction}
\label{chap:intro}

This chapter introduces the motivation behind the thesis, identifies the specific research gap in current literature, and outlines the core objectives and contributions.

\section{Background and Motivation}
\label{sec:intro_bg}

Remote sensing enables quantitative ecological monitoring. Estimating forest attributes like canopy structure (the spatial arrangement of the uppermost branches and leaves) and biomass (the total mass of living plant material) dictates the accuracy of global carbon cycle models \cite{asner2010highres}. Light Detection and Ranging (LiDAR) and Structure-from-Motion (SfM) photogrammetry \cite{schoenberger2016sfm} routinely capture forest topologies. They fail, however, in heavily self-occluded environments. The layered arrangement of dense foliage blocks line-of-sight for both optical and active sensors. This physical occlusion leaves lower canopy layers and underlying terrain entirely unsampled. The result is shadowing. Critical biomass estimates are skewed by these incomplete data representations.

The problem is geometric, not one of sensor quality. In a closed broadleaf canopy only a small part of the sky is visible from the forest floor. A camera ray that enters such a canopy stops at the first leaf it meets. A better sensor does not help here, because the missing information was never carried by any ray that reached it. The only way out is to combine many rays that entered the canopy through different gaps.

Synthetic Aperture Imaging (SAI) does exactly that \cite{kurmi2019statistical}. By aggregating multiple observations from displaced aerial viewpoints, SAI synthetically expands the imaging aperture. This achieves high-resolution capture with a narrow depth of field \cite{youssef2025deepforest}. The system computationally refocuses on specific depth planes. Partial occlusions in the foreground are blurred out, allowing the sensor to effectively look through gaps in the canopy. Multi-perspective reflectance data is merged into a dense volumetric representation. This makes SAI highly effective for structurally complex forests where standard single-view approaches yield only surface measurements.

The idea behind it is simple. A leaf that hides a target from one camera position does not hide it from a position two meters to the side. Averaging multiple views keeps the target sharp and smears the leaf across the image. With the 81 views used here the leaf only adds a weak haze while the target stays in place. Occlusion is therefore not removed but suppressed, and the suppression gets better as the number of views and the width of the aperture grow \cite{kurmi2019statistical}.

Hardware flexibility is another distinct advantage. While LiDAR relies on single-wavelength laser pulses, SAI processes image streams from passive sensors. Low-cost RGB sensors (regular cameras) or off-the-shelf multispectral cameras are sufficient. Capturing broad spectral ranges allows differentiation between vegetation types. Shifts in near-infrared reflectance indicate plant physiology and stress levels---metrics hard to capture with standard LiDAR. A drone with one machine-vision camera flying a grid pattern is enough to produce the input data, which keeps a survey cheap compared to airborne LiDAR.

Despite this capacity for penetration, SAI does not eliminate self-occlusion. Overlapping leaves, twigs, and branches create deep void spaces that no camera angle can observe. During volumetric reconstruction, these void regions can manifest as structured artifacts. The DeepForest framework predicts noisy values at those areas \cite{youssef2025deepforest}. A reconstructed canopy might appear physically continuous where empty space exists. This corrupts downstream pipelines. Vegetation detection, instance segmentation, and subsequent biomass calculations become highly error-prone. Filtering out these void areas would make it possible to utilize SAI data in forestry.

Void regions are not random noise, which makes them a problem. Structures can appear where the integration of many views gives a consistent but wrong reflectance value, so they can look like real matter to any filter that works on intensity alone. A voxel deep inside the canopy that no camera ever saw still gets a value, because the integration averages whatever the contributing rays happened to hit. That value usually sits in the middle of the intensity range, and it changes slowly from voxel to voxel. Both properties are the ones a classical filter uses to decide that something is a real surface.

\begin{figure}[htbp]
    \centering
    \includegraphics[width=0.8\textwidth]{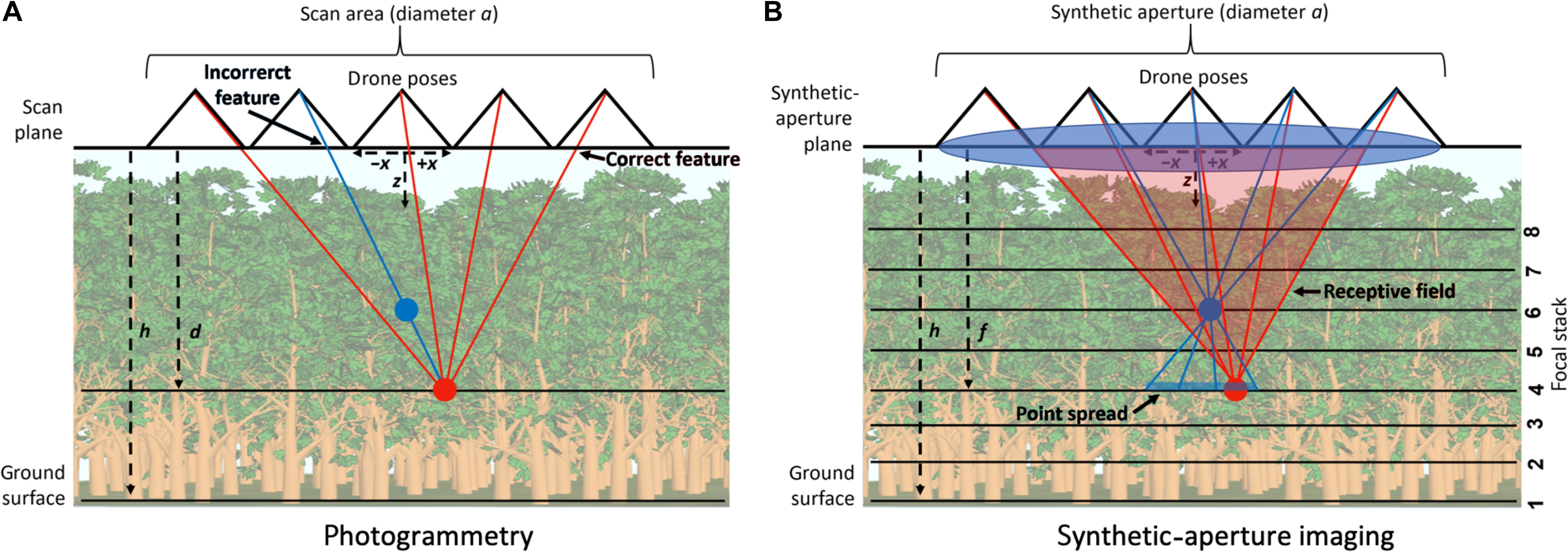}
    \caption{Conceptual overview of the synthetic aperture imaging technique penetrating dense forest canopies, as explored in the DeepForest framework \cite{youssef2025deepforest}.}
    \label{fig:deepforest_concept}
\end{figure}

\section{The DeepForest Context and Identified Gap}
\label{sec:intro_gap}

The DeepForest framework \cite{youssef2025deepforest} established a learning-based baseline for processing dense forest environments from SAI preprocessed aerial imagery. It applies 3D Convolutional Neural Networks generated focal stacks after passing it through the Point-Spread-Function as a transformation. This demonstrated that data-driven interpretation of heavily occluded canopies outperforms deterministic geometric rendering. Natural vegetative scenes are highly irregular. Learned feature extractors handle this self-occluding complexity more robustly than classical filtering techniques.

The goal of the DeepForest network determines the training data. The task is reflectance estimation, and a reflectance target has to be anchored on a voxel that carries a measured value. Empty air carries none, and neither does the inside of a tree trunk. Both are therefore outside the training distribution. This follows from the task. A regression target needs a measured value to regress toward, so dropping the unsampled positions is necessary for a reflectance estimator.

The consequence is that the network is never asked the question this thesis asks, which is whether a given position contains matter at all. At inference time it still returns a reflectance value everywhere, including at the positions it saw no example of, and nothing in the training says what that value should be there. The DeepForest paper pairs the network output with data from SfM techniques partly for this reason, to make the resulting renders easier to read.

For a downstream consumer that matters, because empty canopy gaps still carry values that read as weak returns. Anything that aggregates over the volume then counts them. Estimates for carbon sequestration (the biological process of capturing and long-term storing atmospheric carbon dioxide within plant matter) grow with the number of voxels counted as vegetation, so an error of this kind does not cancel out over a large area. NDVI is more sensitive still, since it is a ratio of the red and near-infrared bands and gets unstable wherever both bands are small. Separating matter from empty air therefore has to happen before any product built on top of the volume can be trusted. That separation is a classification task and not a regression task.

Void labels are not a new idea in this setting. A dual-head variant that predicted reflectance and void points from one network was tried by the DeepForest authors and did not reach a usable result (\autoref{sec:rw_positioning}). The published framework does estimate void points, but only the ones above the top vegetation layer, and it does so through photogrammetric depth reconstruction rather than through the network \cite{youssef2025deepforest}. Void regions below the canopy are named there as an open problem. No accuracy is reported for that filtering step, so there is no learned void result to measure against. What this work proposes is to separate the two. Void estimation becomes a task of its own, with its own labels and its own loss, and no reflectance objective shares the network or influences the architecture.

Two properties of the problem suggest that a supervised formulation should work. First, void space usually has a distinct lack the sharp intensity change that a real structures would produces when the focal plane crosses them. Second, that signature is spread across neighboring depth planes, so it is easier to identify if a system has access to many layers. A convolutional neural network with a receptive field that covers the depth axis can use both. This is the observation the present work builds on.

\section{Thesis Objectives and Contributions}
\label{sec:intro_obj}

This thesis, titled \textit{Predicting Void Points in Self-Occluding Forest Volumes}, introduces a 3D Convolutional Neural Network formulation trained explicitly on data containing zero-reflectance voxels. A semi-realistic synthetic data pipeline leverages Gazebo physics simulations derived from the DeepForest framework \cite{youssef2025deepforest} coupled with NIR-GAN \cite{donike2025nir} to generate paired training data (detailed in \autoref{chap:methodology}). The resulting self-occluded volumetric reflectance stacks provide absolute ground truth for empty space. The network is then optimized to predict and filter out void points from raw SAI focal stacks.

The core component is a custom 3D U-Net configured for volumetric reflectance data. Forest canopy features are highly anisotropic; vertical structures differ drastically from horizontal spread. To preserve critical elevation details during early feature extraction, the multi-layered encoder utilizes asymmetric pooling ( Spatial dimensions $(x, y)$ are downsampled while the depth dimension $(z)$ remains initially intact ). The bottleneck compresses this representation before a symmetric decoder reconstructs the hierarchical features to the original input resolution. A final $1\times1\times1$ convolution reduces the decoder features to a single channel, and a max projection along the depth axis collapses that channel into one probability map for the target plane. Sliding this operation over all 550 focal planes assembles the full occupancy volume, so the network is trained on 2D targets while reasoning over a 3D input window.

\subsection*{Contributions}

The work makes five contributions.

\subsubsection{Void-aware training.}

Zero-reflectance voxels are included as explicit negative labels rather than dropped from the training set. This directly targets the gap described in \autoref{sec:intro_gap}. The label volumes are derived from a separate depth-clipped render of the simulated scene, which makes the ground truth independent of the reflectance stack the network sees as input. Treating void detection as a task of its own, separate from reflectance estimation, is what keeps the formulation simple: one binary target, one loss, and no second objective sharing the network (\autoref{sec:rw_positioning}).

\subsubsection{An updated synthetic data pipeline.}

The Gazebo forest generator is extended with three tree species, photorealistic ground textures, and per-scene random seeds. Rendered RGB is translated to synthetic near-infrared with a pre-trained Pix2Pix generator, and the NDVI index is computed from the red and near-infrared channels. All three channels are then pushed through Light Field Rendering to form the focal stacks (\autoref{sec:data_gen}).

\subsubsection{Architecture study across four input representations.}

Four models are trained and evaluated on the same held-out scene: a 2D U-Net on single planes, a 2D U-Net on stacked planes, a native 3D U-Net on a multi-channel volumetric window, and a transfer-learning variant built on the DeepForest backbone. The comparison isolates what the volumetric receptive field and the spectral channels actually contribute (\autoref{sec:model_arch}).

\subsubsection{A comparison against classical reconstruction.}

COLMAP and DepthAnything3 are voxelized onto the same grid and given a 100-step morphological dilation sweep, so each baseline is reported at its own best operating point instead of at an arbitrary one (\autoref{sec:comparison}).

\subsubsection{non PSF based learning}

The PSF from the DeepForest paper is dropped from the three proposed architectures. In the DeepForest pipeline it was introduced to reduce model size and required compute power without reducing performance. Here the network gets the focal stack directly and chooses the most important features itself, so it can learn the PSF if that is the most meaningful representation of the data. Only the v4 transfer variant keeps it, since it reuses the pre-trained DeepForest backbone and that backbone expects the PSF output as its input (\autoref{sec:dense_unet}).

The primary objective is the classification of void regions in forest volumes. That is the only task this work covers. Reflectance estimation is left to the DeepForest 3D CNN \cite{youssef2025deepforest} and is not replaced here. Marking the empty regions is what isolates the actual vegetation in the volume. Clean volumetric data is an operational requirement for subsequent remote sensing tasks.

Two questions guide the evaluation. Does a network that sees several focal planes at once separate matter from void better than one that sees a single plane? And does a model trained only on simulated forests produce usable output on real drone captures? The first question is answered quantitatively on a held-out synthetic scene. The second can only be answered qualitatively, because no annotated real volumetric ground truth exists for these sites (\autoref{sec:results_real}).

\section{Thesis Outline}
\label{sec:intro_outline}

\Autoref{chap:related} reviews the work this thesis builds on. It covers synthetic aperture imaging for occlusion removal, the classical reconstruction methods used as baselines, encoder--decoder architectures for volumetric segmentation, spectral vegetation indices, and the sim-to-real problem that follows from training on simulated data.

\Autoref{chap:methodology} describes the full pipeline. It starts with the Gazebo simulator configuration and the camera grid, continues through Light Field Rendering, the synthesis of the near-infrared channel, and the construction of the ground truth void volumes, and ends with the four network architectures, the training protocol, the hyperparameter search, and the metrics used throughout.

\Autoref{chap:results} reports the evaluation. The four architectures are compared on the held-out synthetic scene, the effect of the hyperparameter search is isolated, the decision threshold is calibrated, and the selected model is placed against COLMAP and DepthAnything3. A qualitative section covers the transfer to real drone captures.

\Autoref{chap:conclusion} summarizes the findings, states the limitations of the current setup, and lists the extensions that follow most directly from them.

\chapter{Related Work}
\label{chap:related}

This thesis sits between three fields that rarely meet: computational imaging, forest remote sensing, and volumetric deep learning. This chapter reviews each in turn and states what the present work takes from it.

\section{Synthetic Aperture Imaging for Occlusion Removal}
\label{sec:rw_sai}

Synthetic aperture imaging borrows its name and the mathematical basics from radar, where a moving antenna is treated as one large antenna to gain angular resolution \cite{kurmi2019statistical}. The optical version replaces the antenna with a camera and the radar echo with reflected light. A set of images taken from displaced positions is warped onto a common plane and averaged. Anything that lies on that plane stays sharp. Anything in front of or behind it is shifted by a different amount in each view and blurs out.

Kurmi et al.\ give the statistical account of why this removes occlusion \cite{kurmi2019statistical}. Their analysis treats each occluder as a random process that hides a given target point in some fraction of the views. Once the number of views is large enough that at least a few of them see the target, the integrated signal contains a recoverable contribution from it. They measure this with a "visibility" score rather than a signal-to-noise ratio, and the score does not keep growing forever: widening the aperture only helps up to the point where the shift between views matches the size of the occluders, and going wider than that adds nothing. Roughly 50 views are enough to reach 98\% of the achievable visibility, and that number barely changes with how dense the occlusion is. The practical conclusion is that occlusion density limits how much of the target can ever be recovered, not how wide the flight pattern has to be.

Refocusing the same image set at many depths produces a focal stack. Each plane in the stack is a separate integration, and the whole stack is a volume indexed by two image coordinates and one depth coordinate. This is the input representation used throughout this thesis. It is important to be precise about what a voxel in that volume means. It is not a measurement of the material at that 3D position. It is the average of all rays that would pass through that position if the focal plane were placed there, which includes rays that were blocked long before reaching it. A bright voxel therefore means "the views agree at this depth", not "there is matter here".

DeepForest \cite{youssef2025deepforest} applies this to forest canopies and adds a learned stage on top. Instead of picking the in-focus depth per pixel with a hand-written sharpness measure, it trains a 3D CNN on the focal stack and lets the network decide. The paper shows this beats deterministic focus measures on scenes with heavy self-occlusion, which is the result the present work takes as its starting point. Its output is a reflectance estimate rather than an occupancy answer, since the training pairs are anchored on points with a measured reflectance value (\autoref{sec:intro_gap}).

Earlier work on airborne optical sectioning also used thermal instead of visible imaging, on the reasoning that a warm body under a canopy contrasts more strongly against cool foliage than a visible-light target does \cite{schedl2020search}. That line is aimed at search and rescue, where the target is a person and a single detection is enough. Forest structure estimation is a different problem: the target is the canopy itself, it fills the whole volume, and a per-voxel answer is required rather than a single detection. The occlusion-removal machinery carries over, the task formulation does not.

\section{Volumetric Reconstruction of Vegetation}
\label{sec:rw_reconstruction}

\subsection{Structure from Motion and Multi-View Stereo}
\label{sec:rw_sfm}

Structure from Motion recovers camera poses and a sparse 3D point set from an unordered image collection. The classical pipeline detects keypoints, matches them between images, and solves jointly for poses and 3D points. SIFT \cite{lowe2004sift} is still the default detector for this stage because its descriptors survive scale and rotation changes. COLMAP \cite{schoenberger2016sfm} is the standard open implementation of this SfM stage and produces the sparse point set together with the recovered camera poses. COLMAP also ships a separate multi-view stereo stage that can densify this sparse result into a per-pixel depth map per image and fuse it into a dense cloud, though that stage is a distinct piece of software from the incremental SfM method cited here.

The method rests on one assumption: that a small image patch around a keypoint is distinctive enough to be matched to exactly one patch in another image. Forest canopies violate this badly. Iglhaut et al.\ review the use of SfM in forestry and list the failure modes directly \cite{iglhaut2019structure}. Leaves repeat, so a patch on one leaf matches hundreds of others equally well. Foliage moves between exposures, so the same physical point is not in the same place in two images. Occlusion means a point visible in one image is hidden in the next, which breaks the track before it is long enough to be triangulated. Their conclusion is that SfM works well for canopy surfaces and for open stands, and poorly for anything under the canopy.

This matters for the comparison in \autoref{sec:comparison}. The failure of COLMAP on these scenes is not an implementation problem that a better parameter setting would fix. It is the expected behavior on repeating texture with heavy occlusion. What COLMAP does recover tends to be correct, because a match that survives the ratio test and the geometric verification is usually a real correspondence. The result is a small number of accurate points rather than a large number of noisy ones.

\subsection{Monocular Depth Estimation}
\label{sec:rw_mde}

Monocular depth estimators predict a depth or disparity value for every pixel of an image. Modern versions are trained on very large mixed datasets and generalize across scene types without per-scene tuning. DepthAnything3 belongs to this family and is used here as the second baseline \cite{lin2025depthanything3}. It is not restricted to a single image: given a set of views it predicts depth and camera pose jointly for all of them and fuses the result into one point cloud, and that is the mode used in \autoref{sec:comparison}.

The relevant property is what the output actually is. The prediction target stays one depth per pixel, so each view describes a surface, not a volume, and fusing many views stacks surfaces instead of filling a body. Where a real scene has several surfaces along one ray, the estimator has to pick one, and it picks the visible one. For a forest that means the top of the canopy. Everything below is not predicted incorrectly, it is simply not represented in the output format at all. Converting such a depth map to a volume can only produce a shell, or a solid block if the space under the shell is filled in.

These estimators are also scale-ambiguous. They predict depth up to an unknown scale and offset, which has to be fixed against some external reference before the result can be placed in a metric frame. In this work that alignment is done by hand, by fitting the point set into the bounding box of the SAI target volume (\autoref{sec:comparison}).

\subsection{LiDAR and Point Cloud Processing}
\label{sec:rw_lidar}

Airborne LiDAR measures range directly and penetrates canopy gaps better than passive imaging, because a laser pulse can return an echo from several depths along one ray. It remains the reference instrument for forest structure. Its drawbacks are cost, weight, and the fact that a pulse still needs an unobstructed path, so the deepest parts of a closed canopy stay unsampled. Also it usually only operates on a single wavelength, so it misses spectral information required for biological analysis.

Once a point cloud exists, morphological operators are a common way to turn it into a solid model. Gorte and Pfeifer voxelize laser-scanned trees and apply 3D mathematical morphology to recover connected branch structure from the discrete points \cite{gorte2004structuring}. The dilation sweep used in \autoref{sec:comparison} follows the same voxel-based structuring element scheme. Their motivation is the same as the one here: a set of isolated points is not a volume, and comparing it against a solid reference is unfair until it has been grown into one.

\section{Encoder--Decoder Networks for Volumetric Segmentation}
\label{sec:rw_networks}

The U-Net \cite{ronneberger2015u} is the architecture this work is built on. Its shape is an encoder that halves the spatial resolution and doubles the channel count at each level, a bottleneck, and a decoder that mirrors the encoder back up to the input resolution. The part that matters is the skip connection: the feature map from each encoder level is concatenated onto the matching decoder level. Without it the decoder would have to reconstruct fine boundaries from the heavily downsampled bottleneck alone. With it, the decoder gets the high-resolution detail directly and only has to learn how to combine it with the coarse semantic information from below.

Two properties made it the default for segmentation with little training data. It is fully convolutional, so it accepts any input size and shares weights across positions. And the skip connections give short gradient paths to the early layers, which keeps training stable even when the network is deep relative to the size of the dataset.

Çiçek et al.\ extend it to volumes by replacing every 2D operation with its 3D counterpart \cite{cicek20163d}. Convolutions become $3\times3\times3$, pooling becomes $2\times2\times2$, and the feature maps gain a depth axis. The stated motivation is sparse annotation of volumetric medical data, where labeling a few slices is feasible and labeling every voxel is not. The cost is memory. A 3D feature map holds one more dimension than its 2D equivalent at the same spatial resolution, and activations dominate GPU memory during training because they all have to be kept for the backward pass. Which was also a limiting factor in the present work and shaped some design decision in \autoref{sec:model_arch}.

Wang et al.\ address the same cost with a two-stage framework, where a first network operates on a downsampled volume to localize the region of interest and a second refines it at full resolution \cite{wang2018two}. That split is not used here, because the void points to be predicted are distributed through the entire volume rather than concentrated in one region, so there is nothing for a localization stage to narrow down.

Attention gates are a common addition to the volumetric U-Net. The gate learns a per-voxel weight for each skip connection, so the decoder can suppress parts of the encoder feature map that are irrelevant to the current class. This is attractive for the present problem, since most of the volume is uninformative background. It is not used in the architectures evaluated here, and it is listed as an extension in \autoref{sec:future_work}.

Sparse convolutions are the other standard answer to the memory cost. Instead of evaluating the kernel at every position, they store only the non-empty coordinates and evaluate only there. Lee et al.\ discuss the trade-offs of putting 3D spatially sparse networks into practice \cite{lee2021putting}. The gain depends on how empty the volume is, and about 70\,\% of the target volume here is void (\autoref{sec:ground_truth}). That is enough to save something, but far from the sparsity these networks are usually built for.

Anisotropic pooling is the design choice this work does adopt because of the chosen output target (single layer output from 3D input). A focal stack does not have the same meaning along all three axes. The two image axes are sampled at the sensor resolution and are directly comparable. The depth axis is sampled by refocusing and its spacing is set by the chosen focal plane increment. Pooling all three equally throws away depth resolution at the first level, which is where the difference between a real surface and a void is most visible. Keeping the depth axis intact for the first two levels and pooling only the image axes preserves that information into the middle of the encoder. The v3 architecture in \autoref{sec:v3_unet} follows this.

\section{Spectral Vegetation Indices and Near-Infrared Synthesis}
\label{sec:rw_spectral}

Healthy leaves absorb strongly in the red band, because chlorophyll uses that energy for photosynthesis, and reflect strongly in the near-infrared, because the internal cell structure scatters those wavelengths instead of absorbing them~\cite{huang2021ndvi}. The gap between the two bands is therefore large for living foliage and small for soil, bark, rock, or dead material. The Normalized Difference Vegetation Index (NDVI) turns this into a single number, originally introduced as the Band Ratio Parameter in a NASA article \cite{rouse1974monitoring}:

\begin{equation}
    \mathrm{NDVI} = \frac{\mathrm{NIR} - \mathrm{RED}}{\mathrm{NIR} + \mathrm{RED}}
    \label{eq:ndvi}
\end{equation}

where $\mathrm{NIR}$ and $\mathrm{RED}$ are the reflectance values in the two bands. The index falls in $[-1, 1]$. Dense green vegetation sits near the upper end, soil near zero, and water below zero~\cite{huang2021ndvi}. Because it is a ratio, it partly cancels out multiplicative effects like illumination strength, which is why it survives changes in sun angle better than either band alone.

Huang et al.\ review how the index is used in current remote sensing practice and note where it stops being reliable \cite{huang2021ndvi}. It saturates over dense canopies, where adding more leaf area no longer changes the reflectance much. It is sensitive to soil brightness where the canopy is open.

The practical problem for NDVI prediction, which was also the goal of the DeepForest framework, is that the simulator renders visible light only. Getting a near-infrared band out of it requires either extending the renderer with a physical reflectance model per material, or translating the rendered RGB into a synthetic NIR band. The second option was chosen.

A solution to the RGB-to-NIR problem has been studied directly. Pandey et al.\ address the unsupervised case, where aligned RGB and NIR pairs are not available \cite{pandey2020unsupervised}. NIR-GAN \cite{donike2025nir} is the model used in this thesis. It is a Pix2Pix variant trained on Sentinel-2 imagery, where the visible and near-infrared bands are co-registered, which removes the alignment problem that makes paired training hard for ground-level data. It also specifically mentions downstream training as one of the use cases for its output.

One caveat has to be stated plainly. A synthesized NIR band is a prediction, not a measurement and can therefore not contain information that was absent from the RGB input. What it does is re-encode the existing information into a form that matches the input format the downstream network expects at inference time, when it will be fed a real measured NIR band. Its value here is domain alignment, not extra physical content. \autoref{sec:nir_verification} quantifies how close the synthesized band gets to a real one.

\section{Synthetic Training Data and the Sim-to-Real Gap}
\label{sec:rw_simtoreal}

Supervised volumetric segmentation needs per-voxel labels. For a forest interior those labels cannot be obtained by hand. The regions that matter most are exactly the ones no instrument can see, so there is nothing for an annotator to look at. Simulation solves this by construction: the generator knows where every leaf and branch was placed, so a perfect occupancy label is available for free.

The cost is that the network then learns the statistics of the simulator rather than the statistics of a forest. The gap shows up in several places in this specific case. Rendered lighting is uniform and shadow-free compared to real sunlight through moving foliage. Sensor noise in a render is either absent or synthetic. Procedural trees repeat their branching pattern, while real ones do not. Real captures contain wind motion between exposures, specular highlights off wet leaves, and atmospheric haze, none of which the simulator produces.

One family of countermeasures is domain randomization. It varies the simulator parameters widely enough that the real world looks like one more sample from the training distribution; the extended species set and ground textures described in \autoref{sec:simulator_setup} are a mild form of this.

The measure actually used here is simpler and sits at two points in the pipeline. Batch normalization inside the network normalizes activations per channel over the mini-batch, so a constant shift in input brightness is canceled out rather than propagated. On top of that, the input channels are normalized per slice to their own minimum and maximum before they reach the network during inference, which removes the systematic brightness difference between the simulated and the real sensor. None of this addresses structural differences between simulated and real geometry. It only removes the imaging part of the gap, and the results in \autoref{sec:results_real} have to be read with that in mind. How well the spectral part of that imaging gap is closed is checked in \autoref{sec:nir_verification}, where the generator is run on a real RGB capture and its output held against the measured NIR frame of the same site. The structural gap is not measured anywhere in this work.

\section{Positioning of This Work}
\label{sec:rw_positioning}

The basic question of the thesis is: Is it possible to reconstruct 3D volumetric structure (void and occupancy classification) from a focal stack constructed with SAI? A couple of related questions followed from that during the work:

\begin{itemize}
    \item How does the method proposed compare to traditional reconstruction methods used for 3d reconstruction?
    \item Is the NIR channel produced by the NIR-GAN information dense enough to be able to replace other channels?
    \item What information is necessary for such a model to be able to learn the difference between void and occupancy? (Is it possible to learn this from a single focal plane or is a volumetric receptive field necessary?)
\end{itemize}

\autoref{tab:related_positioning} places the three method families against each other on the properties that matter for this task.

\begin{table}[htbp]
    \centering
    \caption[Comparison of reconstruction approaches]{Comparison of reconstruction approaches for self-occluding forest volumes. \enquote{Interior} means the method produces an answer for positions below the outermost visible surface.}
    \label{tab:related_positioning}
    \small
    \begin{tabular}{lcccc}
        \toprule
        \textbf{Approach} & \textbf{Input} & \textbf{Output} & \textbf{Interior} & \textbf{Needs labels} \\
        \midrule
        SfM / MVS \cite{schoenberger2016sfm}   & Multi-view & Sparse points  & Partial & No \\
        Learned depth prediction               & One or more views & Depth map & No & No \\
        LiDAR \cite{gorte2004structuring}      & Active scan & Dense points  & Partial & No \\
        DeepForest \cite{youssef2025deepforest} & Focal stack & Reflectance   & Yes     & Yes \\
        This work                              & Focal stack & Occupancy     & Yes     & Yes \\
        \bottomrule
    \end{tabular}
\end{table}

The two classical baselines share a structural limit. Both derive geometry from visible surfaces, by matching them across views in one case and by predicting a depth per pixel in the other, so both are blind to anything that no camera ray reached. Giving the depth predictor all views instead of one does not lift that limit, it only adds more surfaces. Growing their output with morphology, as done in \autoref{sec:comparison}, can make the result denser but cannot add information that the reconstruction stage never had.

DeepForest \cite{youssef2025deepforest} removes that limit by working on the focal stack, where every position has a value regardless of visibility. What it produces is a reflectance estimate and not an occupancy answer. Its training pairs are anchored on voxels with a measured reflectance value, so empty space carries no label of its own.

The idea of using void labels alongside it has been tried before. The DeepForest paper reports that adding void points to the reflectance training, with a target of zero reflectance at those points, makes the training fail to converge (see chapter `Reconstruction error suppression' of \cite{youssef2025deepforest}). A dual-head variant predicting reflectance and void points at the same time was tried as well and did not reach a usable result.\footnote{Unpublished work by the DeepForest authors, personal communication.} It is therefore not the labels that were missing. What this work changes is that the two tasks are separated, so the void task gets a network of its own and does not compete with a reflectance objective for capacity. This is also why DeepForest is not used as a baseline in \autoref{sec:comparison}. Its void filtering reaches only the space above the top vegetation layer and no accuracy is reported for it, so there is nothing to score against inside the canopy, which is where this work operates. The comparison is made against classical reconstruction methods instead.

The change made here compared to the DeepForest architecture is to the labels and the models, not to the imaging. The ground truth volumes are generated from the same, separate depth-clipped render that marks occupancy directly (\autoref{sec:ground_truth}), which makes empty space a first-class negative class rather than a gap in the training set. Some post processing is applied to the sliced volumes. Everything else in the pipeline follows from that decision: the network is a segmentation model rather than a regression model, the loss is binary cross-entropy over voxels rather than a reflectance error, and the evaluation uses overlap metrics rather than intensity errors.

\chapter{Methodology and Approach}
\label{chap:methodology}

Building upon the objectives outlined in \autoref{chap:intro}, this chapter details the methods used to generate our dataset and the network architectures evaluated. It follows the pipeline in the order the data moves through it: simulation, spectral synthesis, light field rendering, ground truth construction, training, and evaluation.

\section{Training Data Generation}
\label{sec:data_gen}
Real-world volumetric data of forest environments is very expensive to acquire. Annotating it accurately is very complicated due to severe self-occlusion. A synthetic data generation pipeline was implemented to train the models. It simulates the image acquisition process of Synthetic Aperture Imaging (SAI). Then the NIR-GAN is used to enrich the RGB image with an NIR channel \cite{donike2025nir}. This approach (without the NIR-GAN) closely mirrors the methodology of the DeepForest framework \cite{youssef2025deepforest}.

\begin{figure}[htbp]
    \centering
    \begin{tikzpicture}[
    node distance=0.3cm and 0.5cm,
    pstep/.style={
      rectangle, rounded corners=1pt, draw=black!70, fill=blue!6,
      text width=1.4cm, minimum height=0.4cm, align=center,
      font=\tiny
    },
    ch/.style={
      rectangle, rounded corners=1pt, draw=black!70, fill=green!8,
      text width=1.5cm, minimum height=0.4cm, align=center,
      font=\tiny\itshape
    },
    nn/.style={
      rectangle, rounded corners=1pt, draw=black!70, fill=orange!8,
      text width=1.5cm, minimum height=0.4cm, align=center,
      font=\tiny\itshape
    },
    io/.style={
      rectangle, rounded corners=1pt, draw=black!70, fill=gray!10,
      text width=1.5cm, minimum height=0.4cm, align=center,
      font=\tiny\itshape
    },
    arr/.style={->, >=stealth, semithick, black!60},
    bigarr/.style={->, >=stealth, thick, black!50},
  ]

  \node[pstep] (sim) {Simulator};
  \node[ch, right=of sim] (rgb) {RGB};
  \node[nn, below=of rgb] (nirgan) {NIR-GAN \cite{donike2025nir}};
  \node[ch, right=of nirgan] (nir) {NIR};
  \node[ch, right=of rgb] (red) {RED};
  \node[ch, below=of nir] (ndvi) {NDVI};
  \node[pstep, right=of nir] (lfr) {LFR (SAI)};
  \node[pstep, above=of lfr] (aug) {Augmentation};
  \node[pstep, right=of lfr] (tra) {Training};
  \node[nn, below=of tra] (unt) {3D UNet};
  \node[io, below=of ndvi] (rea) {Real Data\\RED, NIR, NDVI};
  \node[pstep, right=of rea] (lfr2) {LFR (SAI)};
  \node[pstep, right=of lfr2] (inf) {Inference};
  \node[io, right=of inf] (voi) {Void Volume};

  \draw[arr] (sim) -- (rgb);
  \draw[arr] (rgb) -- (nirgan);
  \draw[arr] (nirgan) -- (nir);
  \draw[arr] (rgb) -- (red);
  \draw[arr] (red.south) -- (ndvi.north);
  \draw[arr] (nir) -- (ndvi);
  \draw[arr] (red.east) -- ++(0.25,0) |- (lfr);
  \draw[arr] (ndvi.east) -- ++(0.25,0) |- (lfr);
  \draw[arr] (nir.east) -- (lfr);
  \draw[arr] (lfr) -- (aug);
  \draw[arr] (lfr) -- (tra);
  \draw[arr] (aug.east) -- ++(0.25,0) |- (tra);
  \draw[arr] (tra) -- (unt);
  \draw[arr] (unt) -- (inf);
  \draw[arr] (rea) -- (lfr2);
  \draw[arr] (lfr2) -- (inf);
  \draw[arr] (inf) -- (voi);

  \node[ch, right=of nirgan] (nir2) {NIR};

  \begin{scope}[on background layer]
    \node[rectangle, rounded corners=2pt, draw=black!30, fill=gray!3, fit=(sim)(voi)(rea), inner sep=6pt] (allbox) {};
  \end{scope}

  \node[anchor=north west, font=\tiny\bfseries, inner sep=2pt] at (allbox.north west) {v3};

\end{tikzpicture}
    \caption{The pipeline as used for v3, marked accordingly. It shows the data generation from the Gazebo simulator, translation via NIR-GAN, processing into volumetric stacks via LFR, and training the 3D U-Net on the RED, NIR and NDVI channels. \autoref{fig:pipeline_rgb} shows the simpler pipeline v1 and v3b use instead.}
    \label{fig:pipeline}
\end{figure}

Each simulated scene produces two independent image sets. The first is a $9\times9$ grid of top-down captures, taken from 81 displaced positions at a fixed altitude. This set is the input to Light Field Rendering and becomes the focal stack the network sees. The second is a stack of 550 depth-clipped renders taken from a single fixed camera, where the near and far clipping planes are advanced together in small steps so that each render shows only the pixels inside one thin slab. This second set is what becomes the ground truth. The two sets are produced by separate simulation passes over the same scene with the same random seed, so they describe the same forest.

\subsection{Simulator Setup}
\label{sec:simulator_setup}
The training dataset foundation relies on the Gazebo simulator, which was also used in the DeepForest framework \cite{youssef2025deepforest}. The scenes are populated with three different tree species exhibiting realistic leaf arrangements and varying densities (trees per area). The training data can be found in the following Github repository: \url{https://github.com/Ineb01/training_data_self_simulated}.

The forest plugin generates each stand from a small parameter set. The patch is 70 units on a side. Tree counts vary per scene and cycle through a fixed list ranging from 80 to 200 trees, which gives stands from open woodland to closed canopy inside one dataset. The random seed is set to the scene number, so any scene can be regenerated exactly. \autoref{tab:sim_params} lists the values actually used for the \texttt{training-02} dataset.

\begin{table}[htbp]
    \centering
    \caption{Simulator configuration used to generate the dataset on Github (training-02). Tree counts cycle through the listed values as scenes are generated, so the dataset covers a range of canopy densities rather than one fixed density. Mostly the parameters are the same as in the original DeepForest implementation, but the tree variety and ground textures have been diversified.}
    \label{tab:sim_params}
    \small
    \begin{tabular}{lll}
        \toprule
        \textbf{Group} & \textbf{Parameter} & \textbf{Value} \\
        \midrule
        Forest    & Patch size              & $70 \times 70$ units \\
                  & Trees per scene         & 80 to 200 (16-value cycle) \\
                  & Random seed             & Scene number \\
                  & Ground texture index    & 0 \\
                  & Direct spawning         & Enabled \\
        \midrule
        Species   & Birch                   & 40\,\% of trees \\
                  & Oak                     & 35\,\% of trees \\
                  & Ash                     & 25\,\% of trees \\
                  & Homogeneity             & 0.95 (all species) \\
        \midrule
        Tree shape & Branch factor          & 2.45 \\
                  & Levels / segments       & 6 / 6 \\
                  & Trunk length            & 0.5 \\
                  & Taper rate              & 0.8 \\
                  & Twist rate              & 8.0 \\
                  & Clump range             & 0.40 to 0.45 \\
        \midrule
        Top-down grid & Grid resolution     & $9 \times 9$ (81 poses) \\
                  & Baseline extent         & 24 units \\
                  & Pose spacing            & 3 units \\
                  & Camera altitude         & 35 units \\
                  & Camera pitch            & $\pi/2$ (nadir) \\
                  & Near / far clip         & 1 / 35.01 units \\
        \midrule
        Z-stack   & Layers                  & 550 \\
                  & Camera position         & $(0, 0, 35)$ \\
                  & Near clip start         & 34.78 units \\
                  & Far clip start          & 35.05 units \\
                  & Clip step               & 0.03 units \\
                  & Layers per batch        & 10 \\
        \bottomrule
    \end{tabular}
\end{table}

Two details in that table deserve comment. The 24-unit baseline is deliberately much smaller than the 70-unit patch, so all 81 views look at the same central part of the stand and the aperture stays fully populated across the region of interest. And the z-stack uses a slab thickness of $35.05 - 34.78 = 0.27$ units advanced in 0.03-unit steps, so consecutive slabs overlap by a factor of nine. The overlap is intentional and makes the ground truth continuous along depth instead of a set of disconnected shells.

Generating 550 separate simulation runs per scene would be very slow. The z-stack is therefore produced in batches of 10 layers per launch, which cuts the number of Gazebo start-ups by 10 and therefore speeds up overall process time for the simulation by approximately 10 times.

The original DeepForest implementation \cite{youssef2025deepforest} relied on rudimentary procedural trees and bare ground representations. That design choice yielded stable training at the time, but the lack of visual complexity limits model generalization. The updated simulator introduces diverse tree species with complex branch architectures. Added realism forces the network to learn more robust feature representations rather than memorizing low-texture synthetic artifacts.

\begin{figure}[htbp]
    \centering
    \includegraphics[trim={2cm 0 0 0},clip,height=0.27\textwidth]{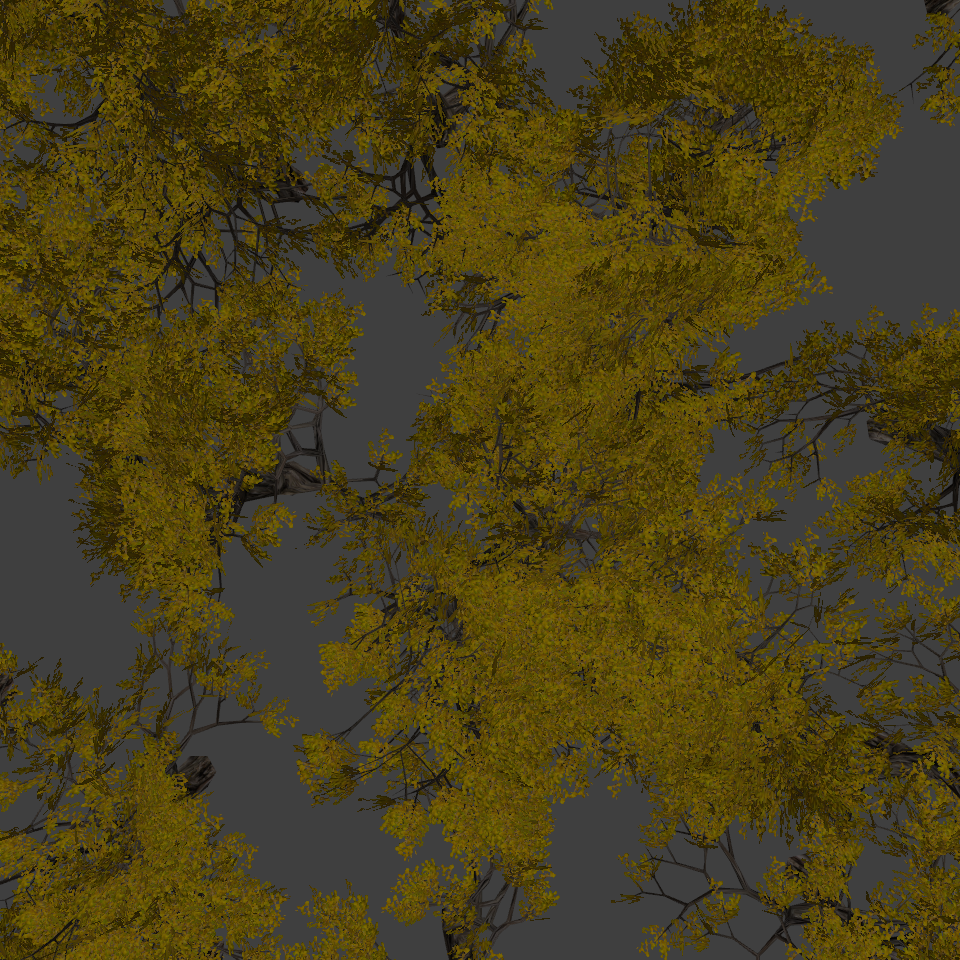}
    \hfill
    \includegraphics[height=0.27\textwidth]{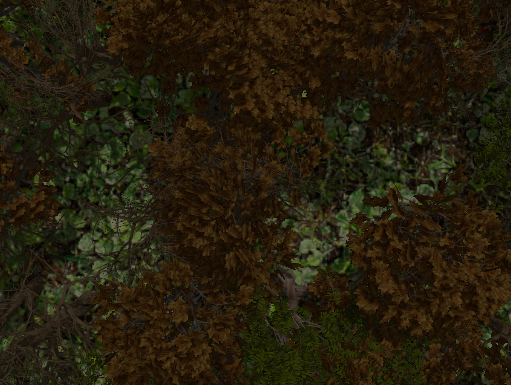}
    \hfill
    \includegraphics[height=0.27\textwidth]{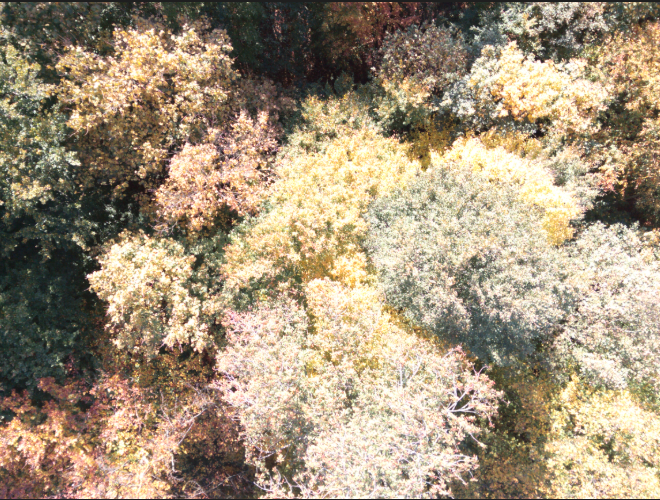}
    \caption{Visual comparison of the data distributions. Left: The simplified DeepForest dataset. Center: The improved simulator with diverse trees and floor textures. Right: Real-world forest data from an October field experiment.}
    \label{fig:simulator_comparison}
\end{figure}

\autoref{fig:simulator_comparison} shows why this matters. The DeepForest scenes have a flat, near-uniform ground plane, so any dark pixel is almost certainly a tree. A network trained on that data can reach low loss by learning a brightness threshold, and such a rule transfers to nothing. The textured ground in the center panel removes that shortcut. The right panel shows the target domain, where ground and canopy overlap heavily in brightness and no threshold separates them.

\subsection{Integration of NIR-GAN}
\label{sec:nir_gan}
Void prediction is a geometric task and needs no near-infrared band to work. NIR is in the pipeline for a different reason. On real captures both the RGB image and the near-infrared band are measured directly, so the choice is not forced by missing sensor data. It is forced by how the two models behave at inference: \autoref{sec:results_real} shows that the variant trained only on the RGB focal stack reaches almost no usable occupancy on a real capture, while the variant reading the RED, NIR and NDVI channels keeps a crown-like shape in two of the three real seasons. The training data therefore has to carry the channels that let the deployed model behave sensibly on real data, even though \autoref{sec:results_v1_discussion} shows this costs some geometric accuracy on synthetic scenes rather than being free.

Gazebo outputs accurate structural ground truths. Its rendered RGB output, which is different from real-world multispectral and Near-Infrared (NIR) imagery. It cannot render NIR images, which biological vegetation analysis relies heavily on NIR because healthy plant cell structures strongly reflect this specific wavelength \cite{huang2021ndvi}.

NIR-GAN \cite{donike2025nir} is integrated into the pipeline to cross this domain gap. It is an image-to-image translation network built upon the Pix2Pix architecture \cite{isola2017image}. It translates simulated visible-spectrum renders into realistic NIR-like images. The plain Pix2Pix configuration is used with the \texttt{S2} checkpoint, which was trained on Sentinel-2 imagery by the authors of the paper and the weights are open-sourced. The optional SatCLIP location conditioning that the NIR-GAN codebase supports is switched off, because it expects geographic coordinates as an additional input and a Gazebo scene has none. Feeding it an arbitrary coordinate would add a signal that carries no information about the scene. The generative model translates the Gazebo outputs without any further fine-tuning.

The generator operates on $512\times512$ patches, while the rendered images are larger. Inference therefore tiles each image with a stride of 448, which gives a 64-pixel overlap between neighboring patches. Overlapping predictions are summed and divided by a per-pixel count of contributions (2 at edges and 4 at corners), so the seams average out instead of appearing as hard edges. Generator output is in $[-1, 1]$ and is mapped to $[0, 1]$ before being written out as 8-bit grayscale.

The red channel needs no synthesis. It is taken directly from the rendered RGB image. NDVI is then computed per pixel from the synthesized NIR and the extracted red band using \autoref{eq:ndvi}, and rescaled from $[-1, 1]$ to $[0, 1]$ so all three channels share one numeric range. Training the 3D segmentation models directly on these GAN-enhanced volumes should harden the learned features against domain shifts.

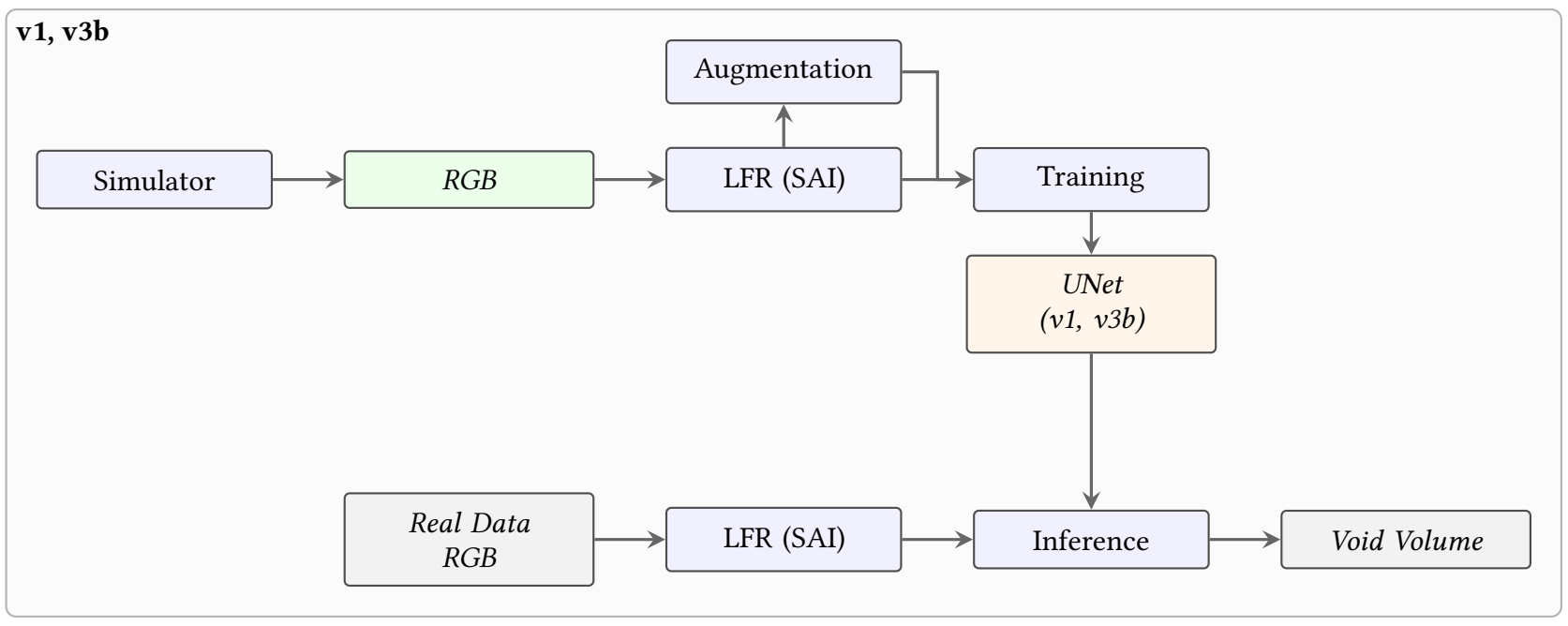
\begin{figure}[htbp]
    \centering
    \begin{tikzpicture}[
    node distance=0.3cm and 0.5cm,
    pstep/.style={
      rectangle, rounded corners=1pt, draw=black!70, fill=blue!6,
      text width=1.4cm, minimum height=0.4cm, align=center,
      font=\tiny
    },
    ch/.style={
      rectangle, rounded corners=1pt, draw=black!70, fill=green!8,
      text width=1.5cm, minimum height=0.4cm, align=center,
      font=\tiny\itshape
    },
    nn/.style={
      rectangle, rounded corners=1pt, draw=black!70, fill=orange!8,
      text width=1.5cm, minimum height=0.4cm, align=center,
      font=\tiny\itshape
    },
    io/.style={
      rectangle, rounded corners=1pt, draw=black!70, fill=gray!10,
      text width=1.5cm, minimum height=0.4cm, align=center,
      font=\tiny\itshape
    },
    arr/.style={->, >=stealth, semithick, black!60},
  ]

  \node[pstep] (sim) {Simulator};
  \node[ch, right=of sim] (rgb) {RGB};
  \node[pstep, right=of rgb] (lfr) {LFR (SAI)};
  \node[pstep, above=of lfr] (aug) {Augmentation};
  \node[pstep, right=of lfr] (tra) {Training};
  \node[nn, below=of tra] (unet) {UNet\\(v1, v3b)};
  \node[io, below=2.0cm of rgb] (rea) {Real Data\\RGB};
  \node[pstep, right=of rea] (lfr2) {LFR (SAI)};
  \node[pstep, right=of lfr2] (inf) {Inference};
  \node[io, right=of inf] (voi) {Void Volume};

  \draw[arr] (sim) -- (rgb);
  \draw[arr] (rgb) -- (lfr);
  \draw[arr] (lfr) -- (aug);
  \draw[arr] (lfr) -- (tra);
  \draw[arr] (aug.east) -- ++(0.25,0) |- (tra);
  \draw[arr] (tra) -- (unet);
  \draw[arr] (unet) -- (inf);
  \draw[arr] (rea) -- (lfr2);
  \draw[arr] (lfr2) -- (inf);
  \draw[arr] (inf) -- (voi);

  \begin{scope}[on background layer]
    \node[rectangle, rounded corners=2pt, draw=black!30, fill=gray!3, fit=(sim)(voi)(rea)(unet)(aug), inner sep=6pt] (allbox) {};
  \end{scope}

  \node[anchor=north west, font=\tiny\bfseries, inner sep=2pt] at (allbox.north west) {v1, v3b};

\end{tikzpicture}
    \caption{The pipeline as used for v1 and v3b. Neither reads a synthesized channel, so NIR-GAN and the red and NDVI extraction steps of \autoref{fig:pipeline} drop out entirely, and training runs straight off the rendered RGB stack. The dashed branch from v3b to inference marks that v3b was only ever scored on synthetic scenes in this thesis; v1 is the only variant of the two that was run on the real captures shown in \autoref{sec:results_real}.}
    \label{fig:pipeline_rgb}
\end{figure}

Not every model needs all of this. v1 and v3b never call NIR-GAN at all, since both are trained directly on the RGB focal stack (\autoref{fig:pipeline_rgb}).

\subsection{Verification of NIR Synthesis}
\label{sec:nir_verification}
\begin{figure}[htbp]
    \centering
    \includegraphics[width=0.9\textwidth]{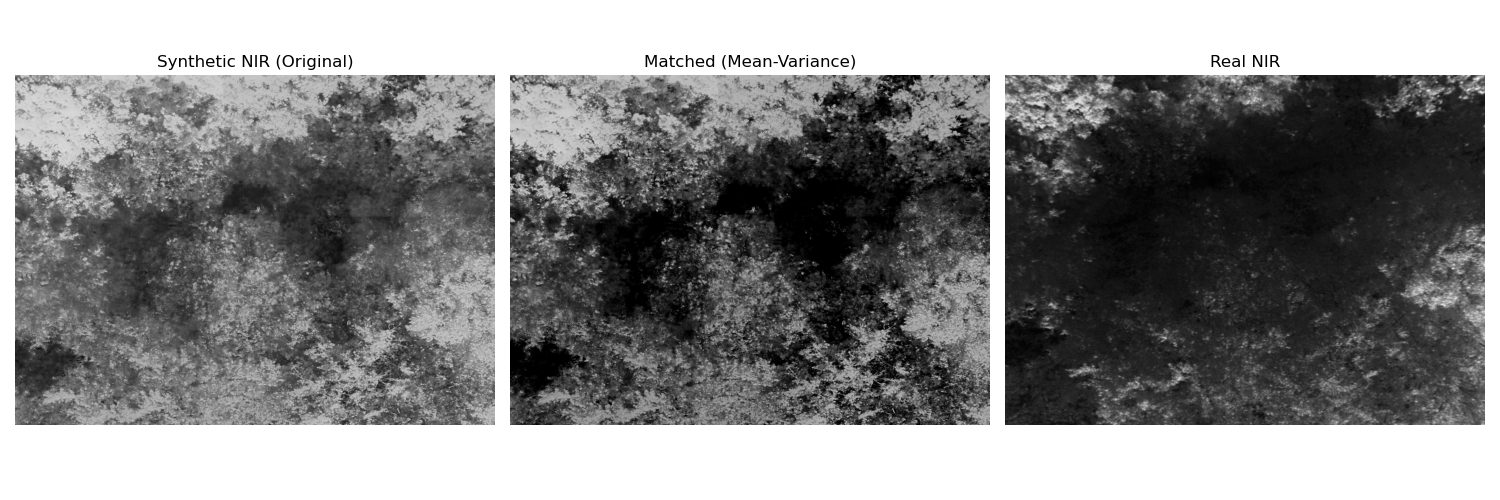}
    \caption{Comparison of the original synthetic NIR (left), the mean-variance matched synthetic NIR (center), and the real-world NIR ground truth (right). Matching the average brightness and its spread aligns the synthetic distribution with the real-world domain.}
    \label{fig:nir_match}
\end{figure}

The synthetic NIR generation fidelity was verified against real-world captures. The downstream 3D U-Net relies on Batch Normalization layers. These layers normalize activations per channel across the mini-batch, making the network inherently robust to absolute shifts in input intensity. What is left is a difference in overall brightness and contrast between the synthetic output and the real sensor data. This is corrected by rescaling the synthetic image so that its average brightness and its spread around that average match the ones measured on real NIR imagery. The synthetic values are first centered on zero and divided by their own spread, then multiplied by the spread of the real data and shifted onto its average:

This post-processing step ensures the structural contrast remains intact while fitting the expected numerical distribution for the 3D CNN (\autoref{fig:nir_match}). Quantitative evaluation shows a Mean Squared Error (MSE) of roughly 0.017 after mean-variance matching. The raw, unscaled generator output yielded an MSE of 0.324.

\begin{equation}
    \hat{x} = \frac{x - \mu_{\mathrm{synth}}}{\sigma_{\mathrm{synth}}} \cdot \sigma_{\mathrm{real}} + \mu_{\mathrm{real}}
    \label{eq:meanvar}
\end{equation}

The size of that drop is the informative part. \autoref{eq:meanvar} is a linear map with two free parameters and cannot change the spatial arrangement of anything in the image. Almost all of the original error was therefore a global offset and scale mismatch, not a structural disagreement, which the generator was placing edges and texture in the right locations from the start and simply predicting the wrong absolute brightness level, which is exactly the kind of error the batch normalization layers downstream should absorb anyway.

What this check does and does not cover should be stated. The generator input here is a real RGB capture from the October field experiment, and the reference is the measured NIR frame of the same site. The MSE is therefore a per-pixel comparison of the same forest, but on a coarse alignment, and part of the remaining 0.017 is that offset rather than generator error.

The more important limit is the input. NIR-GAN is verified on real RGB, while in the pipeline it is fed rendered RGB from Gazebo. Nothing here shows that the translation holds on simulated input, which is the kind of data the training data actually depends on.

\section{Light Field Rendering and Focal Stack Formation}
\label{sec:lfr}

The 81 top-down captures are turned into a focal stack with the Light Field Renderer of the AOS project \cite{kurmi2018airborne}, which is the same that the DeepForest paper uses \cite{youssef2025deepforest}. The specific implementation used here is the LFR module of the AOS reference repository\footnote{\url{https://github.com/JKU-ICG/AOS/tree/stable_release/AOS\%20for\%20Drone\%20Swarms/LFR}}. Each of the 550 output planes is one integration of all 81 views onto a common depth. The internal and external camera parameters are known for the simulator output.

\begin{table}[htbp]
    \centering
    \caption{Light Field Rendering parameters. The same settings are applied to the red, near-infrared, and NDVI channels, so the three resulting stacks are voxel-aligned.}
    \label{tab:lfr_params}
    \small
    \begin{tabular}{ll}
        \toprule
        \textbf{Parameter} & \textbf{Value} \\
        \midrule
        Render resolution      & $512 \times 512$ \\
        Field of view          & $35^{\circ}$ \\
        Input views per plane  & 81 \\
        Focal planes           & 550 \\
        Near focal depth       & $-16.5$ units \\
        Far focal depth        & $+7.8$ units \\
        Plane spacing          & $\approx 0.044$ units \\
        Virtual camera pose    & Center of the $9 \times 9$ grid \\
        \bottomrule
    \end{tabular}
\end{table}

The depth range in \autoref{tab:lfr_params} spans 24.3 units across 550 planes, which places the planes about 0.044 units apart. Which is in the same range as the 0.03 step size of the z-stack in the ground truth, so the input stack and the label stack are sampled at comparable depth resolution and the two can be compared voxel by voxel without resampling along depth.

The rendered result is flipped vertically before it is written out. This is a coordinate format fix and not a processing step. The renderer produces images with the origin at the bottom left, while the image files the rest of the pipeline reads assume the origin at the top left. Getting this wrong would mirror the prediction against the ground truth and destroy any overlap score, which makes it worth stating explicitly.

Each channel is rendered separately with identical parameters. The red stack comes from the extracted red band, the near-infrared stack from the NIR-GAN output, and the NDVI stack from the per-pixel index. Because the geometry and the camera poses are identical across the three runs, the three volumes are aligned voxel for voxel and can be stacked into a single multi-channel tensor without any registration step.

The output naming follows a single format across the whole project. A file named \texttt{TD}$i$\texttt{\_pose\_0\_rgb.png} holds focal plane $i$ of a stack, with $i$ running from 0 at the near depth to 549 at the far depth. Predictions add a \texttt{\_pred} suffix. Every loader in the pipeline reconstructs a volume by reading these files in index order, so a missing or misnamed file shifts the entire depth axis of that volume.

All data from this section can be found precomputed in the following Github repository: \url{https://github.com/Ineb01/training_data_self_simulated.git}

\section{Construction of the Ground Truth Void Volumes}
\label{sec:ground_truth}

The label volumes are an important part of the pipeline that makes the void detection possible. It is also significantly different to the method used in the DeepForest paper so they are described here in full \cite{youssef2025deepforest}.

Ground truth is not derived from the focal stack. It comes from the second simulation pass described in \autoref{sec:simulator_setup}, where a single fixed camera renders 550 depth-clipped images. Any geometry outside that slab is clipped away by the renderer, so a pixel is bright only if a leaf, twig, trunk, or piece of ground actually sits inside the slab at that image position. This gives occupancy directly from the scene geometry, with no inference step in between.

The raw z-stack cannot be used as a label as it is. Three problems have to be fixed first, and a preprocessing pass handles all three.

\subsubsection{Binarization}

The renders carry anti-aliased edges and shading, so a leaf boundary is a gradient rather than a step. Any non-zero pixel is set to 255, which turns the render into a strict occupancy mask. The choice of zero as the cutoff is deliberate. A partially covered pixel still contains matter, and treating it as empty would systematically thin every structure in the label.

\subsubsection{Dilation}

A thin twig seen edge-on inside a 0.27-unit slab covers only a few pixels, and the LFR input cannot resolve structure at that scale. Asking the network to reproduce it would mean asking it to predict something its input does not contain. A circle with a diameter of 11 pixels is applied once to grow the mask to a scale the input can actually support. This is applied per slice, in the image plane only. This is meant to give the learning algorithm a better gradient to follow but is approximately reversible with an erosion operation, so this is actually not a change of target but a change of representation.

\subsubsection{Hole filling}

Dilation grows surfaces and leaves the inside of a trunk or a dense crown hollow. A hollow trunk is a poor occupancy label, since the interior of a solid object actually occupied. The mask is padded by one pixel on each side, flood-filled from the outside corner, and the complement of the filled region is taken as the set of enclosed holes. Those holes are added back to the dilated mask. Padding first is what makes this work: without it, a structure touching the image border would connect to the outside and the flood fill would leak into it.

The result is stored as \texttt{ZS\_processed}, a stack of 550 binary images at $512 \times 512$. Across the full volume of $512 \times 512 \times 550 = 144{,}179{,}200$ voxels the occupied fraction on the held-out scene is 29.6\,\%, so a little under a third. The split is uneven but not extreme. Tree density varies between scenes, so the exact fraction moves across the training set, but the dataset should still be balanced enough to learn from it.

This information is important to interpret the results. A model that labels every voxel as void already reaches 70.4\,\% accuracy on this scene without finding anything, so accuracy on its own says very little here.

The dilation step has one consequence that must be kept in mind when reading the results. It widens every structure in the label by five pixels in each direction. A prediction that recovers the true geometry exactly but does not reproduce the dilation will be counted as under-segmenting. Every score reported in \autoref{chap:results} is therefore a score against the dilated reference, not against the raw simulator geometry. Point clouds should also benefit from this approach because a point can sit five pixels off the true structure in the image plane and still land inside the label. Since all methods are compared against the same reference, the comparison stays fair, but the absolute numbers should not be read as distance to true geometry.

\section{Dataset Composition and Splits}
\label{sec:dataset}

Each scene contributes 550 training samples, one per focal plane. A sample is a small window of the focal stack paired with the ground truth mask of the plane at its center.

\subsection{Channel Normalization}
\label{sec:normalization}

The NIR and RED slices are normalized independently, each to its own minimum and maximum within the slice:

\begin{equation}
    \tilde{x} = \frac{x - \min(x)}{\max(x) - \min(x) + \epsilon}
    \label{eq:minmax}
\end{equation}

This removes any bias that the simulator introduces as it does not render actual reflectance values. In the rendered scenes the synthesized NIR band is consistently brighter than the red band, and a network given both would learn that constant offset as a shortcut for "vegetation" instead of learning a real spectral relationship. Real sensors do not guarantee that offset. Normalizing each band separately destroys it before the network can use it.

NDVI is computed before this normalization, from the raw band values, since \autoref{eq:ndvi} is not meaningful if calculated from individually normalized images. Rescaling it afterwards from $[-1, 1]$ to $[0, 1]$ puts it into the same range as the other two channels.

\subsection{Scene Splits}
\label{sec:splits}

The split is fixed by scene, not by plane. Splitting by plane would put neighboring planes of the same scene on both sides of the split, and since consecutive planes are only 0.044 units apart and the input window spans 16 of them, the training and validation sets would overlap almost completely.

\begin{table}[htbp]
    \centering
    \caption{Dataset split. \texttt{Scene\_201} is excluded from both training and validation and used only for the final evaluation, so no weight and no hyperparameter was ever fitted on it.}
    \label{tab:splits}
    \small
    \begin{tabular}{llr}
        \toprule
        \textbf{Role} & \textbf{Scenes} & \textbf{Samples} \\
        \midrule
        Training   & All remaining \texttt{Scene\_*} & 550 per scene, $\times 8$ augmented \\
        Validation & \texttt{Scene\_202}, \texttt{Scene\_204}, \texttt{Scene\_205} & 1{,}650 \\
        Test       & \texttt{Scene\_201} & 550 \\
        \bottomrule
    \end{tabular}
\end{table}

\texttt{Scene\_201} is held out from training. It appears in neither the training nor the validation list, so it played no part in checkpoint selection or in the hyperparameter search. This is worth stating because the same scene is also used for the baseline comparison in \autoref{sec:comparison}, and the baselines have no notion of a training set at all.

\subsection{Augmentation}
\label{sec:augmentation}

Training samples are augmented eightfold. Four rotations by multiples of $90^{\circ}$, each with and without a horizontal flip. The augmentation index is integrated into the dataset length, so the loader reports eight times as many samples and derives the base sample and the transform from the index. Validation data is not augmented.

Only in-plane transforms are used. Rotating or flipping along the depth axis would be wrong here, because depth is not symmetric: the stack runs from above the canopy down toward the ground, and a vertically mirrored forest is not a plausible forest. Restricting the augmentation to the image plane keeps every augmented sample physically valid.

This is why they are chosen. Multiples of $90^{\circ}$ and axis-aligned flips are exact array operations. This is basically free (computationally, storage and code complexity) multiplication of training data amount. They need no interpolation, so the augmented label stays perfectly binary and the augmented input keeps its original intensity values. An arbitrary rotation would blur both.

\section{Model Architecture}
\label{sec:model_arch}
The primary objective is the prediction of void points within self-occluded volumetric reflectance stacks. This is framed as a 3D semantic segmentation task. Each discrete voxel is classified as either 'matter' or 'void'. Four distinct architectures were evaluated to find the optimal trade-off between representational capacity, spatial awareness, and computational efficiency. The progression shifts from basic 2D frameworks to fully volumetric structures.

The four versions are best explained as experiments of the input representation rather than incremental improvements to performance. Each version changes what the model is allowed to see and holds the rest as close to fixed as the change permits. v1 sees one focal plane in isolation and in RGB. v2 sees several planes, but only as extra channels of a 2D convolution and adds NIR and NDVI channels. v3 sees a real volume, with a kernel that moves along the depth axis. v4 drops the encoder--decoder shape entirely and instead reuses the pre-trained DeepForest backbone on small per-voxel patches. \autoref{tab:arch_summary} summarizes them.

\begin{table}[htbp]
    \centering
    \caption{Overview of the four evaluated architectures. Parameter counts are for the configuration actually trained. The v4 count is given for an axial resolution of 21 input channels, which varies per layer in that model.}
    \label{tab:arch_summary}
    \small
    \begin{tabular}{llllrl}
        \toprule
        \textbf{Version} & \textbf{Type} & \textbf{Input} & \textbf{Depth reasoning} & \textbf{Parameters} & \textbf{Channels} \\
        \midrule
        v1 & 2D U-Net                                                    & $3 \times 448^2$          & None            & 1{,}885{,}441           & RGB \\
        v2 & 2D U-Net, 12 channels                                       & $12 \times 256^2$         & Channel-wise    & 1{,}890{,}625           & RED, NIR, NDVI \\
        v3 & 3D U-Net, from scratch                                      & $3 \times 4 \times 440^2$ & Volumetric      & 5{,}638{,}481           & RED, NIR, NDVI \\
        v4 & Dual-head 3D CNN                                            & $1 \times D \times 2^2$   & Per-voxel patch & 2{,}931{,}362 per layer & GRAY \\
        \bottomrule
    \end{tabular}
\end{table}

\subsection{Baseline U-Net (v1)}
\label{sec:baseline_unet}
The v1 architecture acts as a standard baseline. It is based on the original 2D U-Net \cite{ronneberger2015u}. A pre-trained VGG19 backbone \cite{simonyan2014very} serves as the encoder. Only the first convolutional block is extracted, which is two $3 \times 3$ convolutions with 64 output channels and their ReLU activations. The remaining encoder levels, the bottleneck, and the entire decoder are defined from scratch and trained from random initialization.

The encoder runs at three levels: 64 channels at full resolution, 128 after the first pooling, and a 256-channel bottleneck after the second. The decoder mirrors this with bilinear upsampling followed by a concatenation with the matching encoder feature map, and a $1 \times 1$ convolution with a sigmoid produces the output mask. Batch normalization follows every convolution except the two taken from VGG19, which are used exactly as pre-trained. VGGT Parameters were not frozen for the training step.

This single-plane 2D U-Net operates independently on every layer in the focal stack. Standard 2D convolutions predict the sharpness of local areas within isolated slices. Context is not shared across the depth dimension. This limits its ability to resolve vertical occlusions in principle, though \autoref{sec:results_synth} shows the practical picture is more complicated and its performance was very comparable to the models with access to depth information.

\subsection{Multi-Channel U-Net (v2)}
\label{sec:v2_unet}
\label{sec:res_unet}
To force depth reasoning, the v2 architecture modifies the baseline to accept multiple input channels simultaneously. The input tensor expands from a standard 3-channel slice to 12 channels. The first convolutional layer of the VGG19 backbone was explicitly rewritten. Pre-trained weights were duplicated across the new channel dimension to accommodate the structural change, and divided by the repeat factor of four so that the expected activation magnitude of the layer stays unchanged. Without that division the first layer would produce activations four times larger than the pre-trained statistics of the following layers expect. Also the VGG19 encoder layer was trained on RGB images, so feeding it RED, NIR and NDVI values is a domain shift. For this reason also in this model the weights were also not frozen during training and this will adapt over the training process.

Everything after the first layer is identical to v1. The model therefore isolates one variable: whether giving a 2D network access to several depth planes at once is enough, or whether the convolution itself has to move along the depth axis. The answer matters, because channel-wise access is much cheaper than a true 3D kernel.

The limitation is that a 2D convolution treats its input channels as an unordered set in the pooling step. It learns one weight per channel per output feature and sums them. Depth ordering carries no meaning to such a layer: swapping planes 3 and 9 in the input would give a differently-weighted but equally learnable problem. A 3D convolution cannot do this, because its kernel spans the depth axis and its weights are tied to relative depth offsets.

\subsection{True 3D U-Net (v3)}
\label{sec:v3_unet}
The v3 architecture implements a native 3D U-Net \cite{cicek20163d}. It transitions fully into the 3D spatial domain by utilizing  \texttt{Conv3d} layers, 3D batch normalization, and 3D max pooling. It is a plain encoder--decoder with skip connections, and its behavior comes from the volumetric receptive field and the spectral input.

Each encoder level is a block of two $3 \times 3 \times 3$ convolutions, each followed by 3D batch normalization and a ReLU. The channel count doubles at every level. The decoder mirrors this with transposed convolutions, followed by a concatenation with the skip feature map and another double convolution block. Where a transposed convolution produces a shape that does not match its skip partner exactly, trilinear interpolation corrects it before the concatenation. The input vector is a 4 layer section consisting of RED, NIR and NDVI channels of the layers with offset $-8$, $-4$, $+1$ and $+6$ from the target layer. So the model has access to context around the target layer up and down. The four planes are not adjacent. They are sampled from a window of 16 planes, which reaches further along depth than four neighboring planes would at the same memory cost.

The pooling is anisotropic. The first two levels pool with a $(1, 2, 2)$ kernel, which halves the two image axes and leaves the depth axis untouched. Only the third and any deeper level use a full $(2, 2, 2)$ kernel. With an input depth of four planes this means the depth axis survives intact through the first two levels, the following layers always half this resolution. Pooling depth at the first level would merge adjacent planes and destroy depth resolution evidence.

The synthesized Near-Infrared (NIR) channel is incorporated here. Together with the red band and the derived NDVI index it forms the three input channels. These biological markers could help separate foliage from background.

The output stage differs from a textbook 3D U-Net and is worth stating precisely. A $1 \times 1 \times 1$ convolution reduces the final decoder features to a single channel, giving a volume of shape $(1, D, H, W)$. A max operation along the depth axis then collapses this to $(1, H, W)$, and a sigmoid maps it to probabilities. The network therefore takes a volume and returns a plane. Taking the maximum rather than the mean makes this a detection instead of an average: if any depth in the window supports occupancy at an image position, the output reports it.

The full occupancy volume is assembled at inference time by sliding the target index over all 550 planes and stacking the resulting maps. Each plane is predicted from the four samples of its own 16-plane window, so neighboring output planes come from heavily overlapping inputs.

The cost of that pass is minimal. The selected configuration from the hyperparameter search holds 5{,}638{,}481 parameters, and a single forward pass over one $3 \times 4 \times 440 \times 440$ window takes 66\,ms and peaks at 323\,MiB of VRAM. Assembling the complete 550-plane volume for \texttt{Scene\_201} takes 88 seconds on the hardware used throughout this work, an AMD Ryzen 5 2600 with six cores and 16\,GiB of DDR4-2133 system memory, paired with an NVIDIA GeForce GTX 1070 Ti holding 8\,GiB of VRAM. Only 36 of those seconds are spent inside the network. Decoding, resizing, and normalizing the PNG files each window needs accounts for 47 seconds, and writing the 550 output maps for another 6. Planes are processed one at a time with no batching, and consecutive windows share twelve of their sixteen planes, so the same file is decoded repeatedly. What matters for the argument made here is the order of magnitude. A full volume costs under two minutes on consumer hardware, which means inference is not the part of the pipeline that limits the use of the method.

This architecture is trained entirely from scratch. No pre-trained weights are used, since no large-scale pre-trained 3D encoder exists for volumetric reflectance data and the ImageNet statistics that help v1 and v2 do not transfer to a $3 \times 3 \times 3$ kernel.

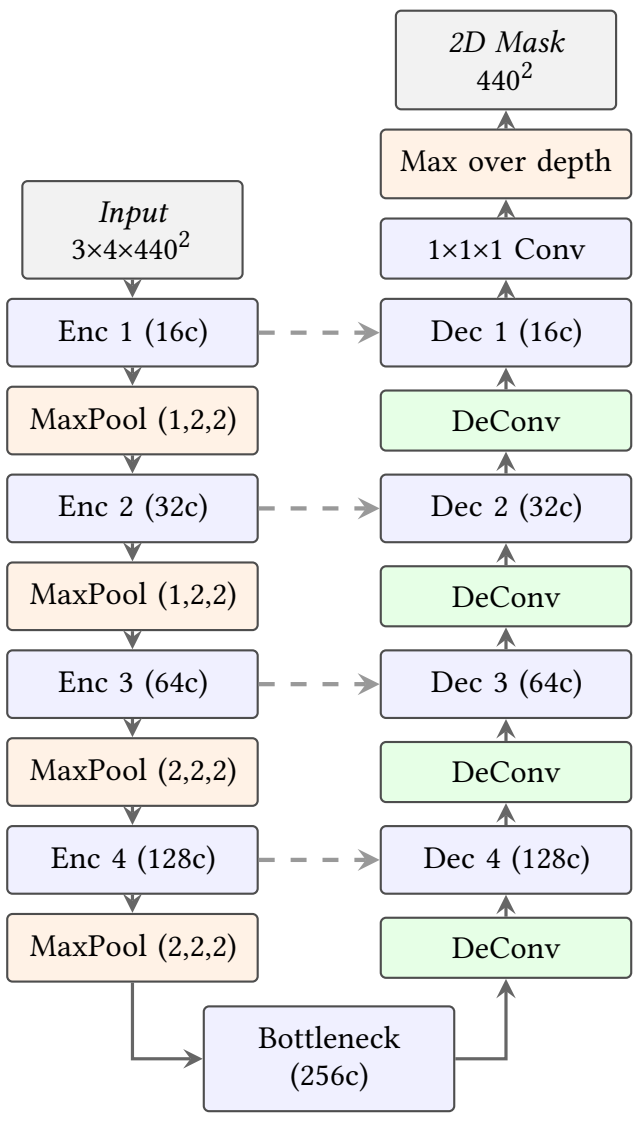
\begin{figure}[htbp]
    \centering
    \begin{tikzpicture}[
    node distance=0.12cm and 0.8cm,
    block/.style={
      rectangle, rounded corners=1pt, draw=black!70, fill=blue!6,
      text width=1.4cm, minimum height=0.22cm, align=center,
      font=\tiny
    },
    pool/.style={
      rectangle, rounded corners=1pt, draw=black!70, fill=orange!10,
      text width=1.4cm, minimum height=0.12cm, align=center,
      font=\tiny
    },
    up/.style={
      rectangle, rounded corners=1pt, draw=black!70, fill=green!10,
      text width=1.4cm, minimum height=0.12cm, align=center,
      font=\tiny
    },
    io/.style={
      rectangle, rounded corners=1pt, draw=black!70, fill=gray!10,
      text width=1.2cm, minimum height=0.18cm, align=center,
      font=\tiny\itshape
    },
    arr/.style={->, >=stealth, semithick, black!60},
    skip/.style={->, >=stealth, dashed, thick, black!40},
  ]

  \node[io] (in) {Input\\$3{\times}4{\times}440^2$};
  \node[block, below=of in] (e1) {Enc 1 (16c)};
  \node[pool, below=of e1] (p1) {MaxPool (1,2,2)};
  \node[block, below=of p1] (e2) {Enc 2 (32c)};
  \node[pool, below=of e2] (p2) {MaxPool (1,2,2)};
  \node[block, below=of p2] (e3) {Enc 3 (64c)};
  \node[pool, below=of e3] (p3) {MaxPool (2,2,2)};
  \node[block, below=of p3] (e4) {Enc 4 (128c)};
  \node[pool, below=of e4] (p4) {MaxPool (2,2,2)};

  \node[block, below=0.15cm of p4, xshift=1.3cm] (bn) {Bottleneck (256c)};

  \node[up, right=0.8cm of p4] (u4) {DeConv};
  \node[block, right=0.8cm of e4] (d4) {Dec 4 (128c)};
  \node[up, right=0.8cm of p3] (u3) {DeConv};
  \node[block, right=0.8cm of e3] (d3) {Dec 3 (64c)};
  \node[up, right=0.8cm of p2] (u2) {DeConv};
  \node[block, right=0.8cm of e2] (d2) {Dec 2 (32c)};
  \node[up, right=0.8cm of p1] (u1) {DeConv};
  \node[block, right=0.8cm of e1] (d1) {Dec 1 (16c)};

  \node[block, above=of d1] (conv) {$1{\times}1{\times}1$ Conv};
  \node[pool, above=of conv] (mp) {Max over depth};
  \node[io, above=of mp] (out) {2D Mask\\$440^2$};

  \draw[arr] (in) -- (e1); \draw[arr] (e1) -- (p1);
  \draw[arr] (p1) -- (e2); \draw[arr] (e2) -- (p2);
  \draw[arr] (p2) -- (e3); \draw[arr] (e3) -- (p3);
  \draw[arr] (p3) -- (e4); \draw[arr] (e4) -- (p4);

  \draw[arr] (p4.south) |- (bn.west);
  \draw[arr] (bn.east) -| (u4.south);

  \draw[arr] (u4) -- (d4); \draw[arr] (d4) -- (u3);
  \draw[arr] (u3) -- (d3); \draw[arr] (d3) -- (u2);
  \draw[arr] (u2) -- (d2); \draw[arr] (d2) -- (u1);
  \draw[arr] (u1) -- (d1); \draw[arr] (d1) -- (conv);
  \draw[arr] (conv) -- (mp); \draw[arr] (mp) -- (out);

  \draw[skip] (e4) -- (d4); \draw[skip] (e3) -- (d3);
  \draw[skip] (e2) -- (d2); \draw[skip] (e1) -- (d1);

\end{tikzpicture}
    \caption{Architecture of the proposed 3D U-Net (v3), shown in the deployed configuration with four encoder levels starting at 16 features. The first two pooling stages use a $(1,2,2)$ kernel and leave the depth axis intact. The final max projection along depth collapses the output volume into a single probability map for the target plane.}
    \label{fig:architecture}
\end{figure}

\subsection{Dual-Head 3D CNN (v4)}
\label{sec:dense_unet}
The v4 architecture directly adopts the model structure and pre-trained weights from the DeepForest project \cite{youssef2025deepforest}. It is not an encoder--decoder network. Three 3D convolution blocks with instance normalization and GELU activations process a tiny input patch of shape $(1, D, 2, 2)$ (this is the output of the PSF), the flattened result is passed to fully connected layers, and a sigmoid produces one scalar per patch. One forward pass therefore predicts one voxel, not one plane. The 3DCNN of the DeepForest framework uses 440x440 images as input so this is what is converted to before using the PSF.

The input data undergoes rigorous preprocessing via a point spread function to match the original DeepForest training distribution. The focal stack is center-cropped to $440 \times 440$, reduced to the red band, and the per-voxel slice contributions are extracted and tiled into the $2 \times 2$ spatial footprint the backbone expects. The axial resolution $D$ is not constant. It is read from a per-layer table and changes with depth, so the fully connected head has to be sized per layer.

A frozen-backbone training paradigm is utilized. The three convolution blocks and the original prediction head are frozen, and a second head of two fully connected layers is attached to the same flattened features and trained on the void prediction target. The generalized features extracted by the DeepForest backbone are then compared directly against the custom-trained 3D U-Net.

The per-voxel formulation is what limits this version in practice. A full plane at $440 \times 440$ needs 193{,}600 forward passes. v4 also inherits the depth convention of the DeepForest backbone, whose per-layer table covers 440 layers rather than the 550 planes the other versions predict, so a full volume needs 440 times that. On the available hardware one plane takes about ten hours of inference. Training was therefore restricted to a single depth layer and ran for roughly 24 hours. No full-volume result exists for v4, and \autoref{sec:results_v4} reports only what was measured on that one layer.

The training data of the DeepForest 3DCNN consists only of occupied voxels, so the model has no void data in its trained domain. This is in itself not a problem for the head, since it can be trained to detect out-of-distribution features if the backbone is unsure. In theory this second head can be interpreted as an out-of-distribution classifier for the backbone which is solely trained on occupied voxels.
\section{Training Protocol and Hyperparameter Optimization}
\label{sec:training}

\subsection{Objective and Optimizer}
\label{sec:objective}

Void prediction is a per-voxel binary problem, so the loss function is binary cross-entropy over the output plane:

\begin{equation}
    \mathcal{L} = -\frac{1}{N}\sum_{i=1}^{N} \left[ y_i \log \hat{y}_i + (1 - y_i) \log (1 - \hat{y}_i) \right]
    \label{eq:bce}
\end{equation}

where $\hat{y}_i$ is the predicted probability for voxel $i$ and $y_i \in \{0, 1\}$ its label. No class weighting is applied. The occupied class covers 29.6\,\% of the held-out volume (\autoref{sec:ground_truth}), so the split is uneven, but nowhere near the ratios that make a weighted loss necessary.

The loss is computed on the sigmoid output. Adam is used as the optimizer, with the default momentum terms in the pytorch implementation.

Checkpointing is driven by validation loss, not training loss. After every epoch the model is evaluated on the three validation scenes and the weights are written out only if the validation loss improved.

\subsection{Hyperparameter Search}
\label{sec:hpo}

The v3 configuration was selected by an Optuna study rather than by hand. The search minimizes validation loss over four parameters, listed in \autoref{tab:hpo_space}.

\begin{table}[htbp]
    \centering
    \caption{Hyperparameter search space for the v3 architecture. Fixed settings are given for reference. The search ran 20 trials, each for four epochs on a deterministic 10\,\% subsample of the training data.}
    \label{tab:hpo_space}
    \small
    \begin{tabular}{lll}
        \toprule
        \textbf{Parameter} & \textbf{Type} & \textbf{Range or value} \\
        \midrule
        Batch size       & Categorical & $\{1, 2, 4\}$ \\
        Learning rate    & Log-uniform & $10^{-5}$ to $10^{-3}$ \\
        Initial features & Categorical & $\{16, 32\}$ \\
        Encoder depth    & Integer     & 2 to 4 \\
        \midrule
        Input size       & Fixed       & $440 \times 440$ \\
        Focal planes     & Fixed       & 4 of a 16-plane window \\
        Epochs per trial & Fixed       & 4 \\
        Subsample        & Fixed       & 10\,\% (seed 42) \\
        \bottomrule
    \end{tabular}
\end{table}

Two decisions kept the search efficient. Trials ran on a deterministic 10\,\% subsample of the training set, drawn with a fixed seed so every trial sees the same subset and the comparison between trials stays fair. And trials do not write checkpoints, since only the validation loss is needed to rank them. Even so the study took several days. Individual trials ranged from about 1.5 hours to nearly 14 hours depending on the sampled configuration.

The upper bounds of the search space were set by GPU memory. A 3D U-Net keeps every activation for the backward pass, and at $440 \times 440$ with four depth planes the activation memory grows quickly with both batch size and feature width. Configurations beyond batch size 4 and 32 initial features exhausted the available memory in early exploratory runs, so the search space was bounded below that point and an \texttt{OutOfMemoryError} handler was left in place as a safety net. Within the bounded space no trial hit it, all 20 trials completed.

\subsection{Selected Configuration}
\label{sec:selected_config}

\autoref{tab:hpo_best} lists the best trials. The winner uses 16 initial features with four encoder levels, which gives a 256-channel bottleneck through the progression $16 \rightarrow 32 \rightarrow 64 \rightarrow 128 \rightarrow 256$.

\begin{table}[htbp]
    \centering
    \caption{The five best trials of the v3 hyperparameter study, ranked by validation loss. Trial 16 was selected. The bottleneck width is the derived quantity $f \cdot 2^{d}$.}
    \label{tab:hpo_best}
    \small
    \begin{tabular}{rrrrrrr}
        \toprule
        \textbf{Trial} & \textbf{Val.\ loss} & \textbf{Batch} & \textbf{Features} & \textbf{Depth} & \textbf{Bottleneck} & \textbf{Learning rate} \\
        \midrule
        16 & 0.3382 & 4 & 16 & 4 & 256 & $3.05 \times 10^{-4}$ \\
        13 & 0.3405 & 4 & 16 & 4 & 256 & $2.06 \times 10^{-4}$ \\
        7  & 0.3426 & 4 & 16 & 3 & 128 & $3.21 \times 10^{-4}$ \\
        18 & 0.3396 & 2 & 32 & 4 & 512 & $4.60 \times 10^{-5}$ \\
        0  & 0.3502 & 4 & 16 & 4 & 256 & $8.27 \times 10^{-5}$ \\
        \bottomrule
    \end{tabular}
\end{table}

The pattern across the study is clearer than the winner alone. Every trial with batch size 1 landed above 0.56 validation loss, far worse than anything at batch size 2 or 4. That is assumed to be a batch normalization effect rather than an optimization effect: with a single sample the batch statistics are computed from one volume and are too noisy to normalize with. Learning rates below $10^{-5}$ also did poorly, but for the ordinary reason that four epochs are not enough to converge at that rate.

Depth and width did not both help. They traded off against each other. Going from three to four encoder levels at 16 features improved the loss, while going to 32 features at four levels did not improve on 16 features at four levels despite doubling the bottleneck to 512 channels. The useful quantity appears to be how far the receptive field reaches, not how many channels the bottleneck holds.

The selected model was then retrained on the full training set, not on the 10\,\% subsample used during the search. \autoref{sec:results_hpo} compares it against the pre-search configuration to isolate what the search actually bought.

\section{Evaluation Metrics}
\label{sec:metrics}

Every metric in \autoref{chap:results} comes from one confusion matrix accumulated over the whole test volume. Predictions and ground truth are compared voxel by voxel, and the counts of true positives, true negatives, false positives, and false negatives are summed across all 550 planes before any ratio is computed.

Accumulating first and dividing afterwards is not the same as averaging per-plane scores, and the difference matters here. Planes near the top and bottom of the stack contain very little matter, so a per-plane IoU on those planes is either undefined or dominated by a handful of voxels. Averaging such values would give those nearly empty planes the same weight as a plane through the middle of the canopy. Summing the confusion matrix first weights every voxel equally instead.

The reported quantities are defined as follows, with $\mathrm{TP}$, $\mathrm{TN}$, $\mathrm{FP}$, and $\mathrm{FN}$ the accumulated counts:

\begin{align}
    \text{Precision} &= \frac{\mathrm{TP}}{\mathrm{TP} + \mathrm{FP}} &
    \text{Recall} &= \frac{\mathrm{TP}}{\mathrm{TP} + \mathrm{FN}} \label{eq:prec_rec} \\[0.4em]
    \text{Specificity} &= \frac{\mathrm{TN}}{\mathrm{TN} + \mathrm{FP}} &
    \text{Accuracy} &= \frac{\mathrm{TP} + \mathrm{TN}}{\mathrm{TP} + \mathrm{TN} + \mathrm{FP} + \mathrm{FN}} \label{eq:spec_acc} \\[0.4em]
    \mathrm{IoU} &= \frac{\mathrm{TP}}{\mathrm{TP} + \mathrm{FP} + \mathrm{FN}} &
    \text{Dice} &= \frac{2\,\mathrm{TP}}{2\,\mathrm{TP} + \mathrm{FP} + \mathrm{FN}} \label{eq:iou_dice} \\[0.4em]
    \mathrm{Balanced Accuracy} &= \frac{\mathrm{Recall} + \mathrm{Specificity}}{2} & \label{eq:balanced_acc}
\end{align}

The Jaccard Index (IoU) and the S\o{}rensen--Dice coefficient are the two headline numbers. Both ignore true negatives, which is what makes them useful for this task. About 70\,\% of the test volume is empty (\autoref{sec:ground_truth}), so a model that labels empty space correctly and never finds a single structure still scores well on accuracy. IoU and Dice give it a low score. The two metrics are related through $\mathrm{Dice} = 2\,\mathrm{IoU} / (1 + \mathrm{IoU})$, so they rank methods the same way. Both are reported because the literature is split on which to quote, and because Dice is less harsh on small overlaps.

For a binary problem the Dice coefficient and the F1 score are the same quantity written two different ways. The results tables therefore show identical values in the F1 and Dice columns.

Balanced accuracy (\autoref{eq:balanced_acc}) is reported alongside plain accuracy. Plain accuracy is pulled up by the empty majority class, balanced accuracy is not, so the balanced version is the one which is more representable here and on any scene with an even sparser canopy.

\subsection{Thresholds and Resampling}
\label{sec:thresholds_resampling}

Two conventions affect the numbers and have to be fixed explicitly.

Predictions are written to disk as 8-bit grayscale, so a probability of $p$ is stored as $\mathrm{round}(255p)$. A decision threshold of 0.5 therefore corresponds to a stored value of 128, and every threshold quoted in this thesis is given in both scales. Ground truth is thresholded at any non-zero value, matching the binarization in \autoref{sec:ground_truth}.

Predictions are produced at $440 \times 440$ while the ground truth is stored at $512 \times 512$, so one resampling step is unavoidable.
\section{Post-Processing and Threshold Optimization}
\label{sec:postprocessing}

To effectively compare the soft, continuous volumetric predictions from the network with the binary ground truth volumes, a binarization step is required. The model outputs probabilities for each voxel, representing the likelihood of belonging to the target object. Rather than assuming a standard cutoff of 0.5, a threshold optimization was implemented to find the theoretical maximum IoU value.

The procedure is a plain sweep. The full predicted volume of 550 planes is loaded once and kept in memory as an 8-bit array. 48 cutoff values are placed evenly between 10 and 245 on the 8-bit scale, which is a step of 5 and covers probabilities from about 0.04 to 0.96. For each candidate the volume is binarized, and IoU and Dice are computed against the ground truth using \autoref{eq:iou_dice}. The cutoff that maximizes the overlap is reported.

Loading the volume once and re-thresholding it in place makes the sweep computationally cheap. Each candidate costs one comparison and two boolean reductions over 144 million voxels, so the whole sweep runs in the time a single re-inference would take.

Two properties of this procedure need to be stated honestly. The sweep is run on the same held-out scene that the result is reported on, so the reported optimum is an upper bound rather than a value that transfers to an unseen scene. And because IoU and Dice are monotonically related, both are maximized by the same cutoff, so reporting two metrics does not provide two independent confirmations.

The optimization is applied to the v3 predictions on \texttt{Scene\_201}. The global maximum sits at a raw intensity of 75, which is about 0.29 on the normalized probability scale. At that cutoff the v3 architecture reaches an IoU of 0.6080 and a Dice coefficient of 0.7562. The default cutoff of 0.5, stored as 128, gives a noticeably lower IoU of 0.5705 and Dice of 0.7265. The gain is 3.75 percentage points of IoU.

\chapter{Results and Evaluation}
\label{chap:results}

This chapter presents the quantitative and qualitative outcomes of the proposed architectures. The quantitative evaluation is carried out on the held-out synthetic scene \texttt{Scene\_201}, which no model saw during training or validation. The transfer to real drone captures is assessed qualitatively, since no annotated real volumetric ground truth exists.

\section{Performance on Synthetic Data}
\label{sec:results_synth}

Following the methodology outlined in \autoref{chap:methodology}, the model variations described in \autoref{sec:model_arch} were evaluated for their ability to detect void points in highly self-occluding forest canopies (the dense upper layer of leaves and branches in a forest). A voxel-level evaluation was carried out on three held-out scenes, described below. The raw predictions were thresholded at 128 on the 0--255 uint8 scale, which corresponds to a probability of 0.5. Ground truth was thresholded at any non-zero value and post processed as explained in \autoref{sec:postprocessing}. All 550 planes of each scene were evaluated, and the confusion matrix was accumulated over the full volume of 144{,}179{,}200 voxels per scene before any ratio was computed.

\subsection{Generalization Across Unseen Densities}
\label{sec:results_density}

Three scenes were built and held out from training, validation and the hyperparameter search, each with an independent forest seed so the tree placement does not overlap between scenes. \texttt{Scene\_301}, \texttt{302} and \texttt{303} carry 150, 175 and 200 trees, which puts their ground-truth occupied fraction at 0.326, 0.366 and 0.462. \autoref{tab:density_sweep} reports voxel-level IoU, per scene and pooled, for the three architectures that completed volume-wide inference plus the pre-parameter-search version of v3 discussed in \autoref{sec:results_hpo}.

\begin{table}[htbp]
    \centering
    \caption[Voxel-level IoU on three unseen density-sweep scenes]{Voxel-level IoU on three scenes excluded from training, validation and the hyperparameter search, at 150, 175 and 200 trees (ground-truth occupied fraction 0.326, 0.366 and 0.462). Pooled accumulates the confusion matrix over all three scenes before computing IoU. All 550 planes were evaluated per scene.}
    \label{tab:density_sweep}
    \small
    \begin{tabular}{lrrrrr}
        \toprule
        \textbf{Model} & \textbf{301 (150)} & \textbf{302 (175)} & \textbf{303 (200)} & \textbf{Pooled} \\
        \midrule
        v1 (2D U-Net)          & 0.6033 & 0.6069 & 0.5903 & 0.5993 \\
        v2 (multi-channel)     & 0.5632 & 0.5710 & 0.6145 & 0.5857 \\
        v3 (3D U-Net)          & 0.5669 & 0.5522 & 0.5883 & 0.5706 \\
        v3 before search       & 0.5407 & 0.5362 & 0.5597 & 0.5466 \\
        v3b (RGB)              & \textbf{0.6528} & \textbf{0.6540} & \textbf{0.6874} & \textbf{0.6668}  \\
        \bottomrule
    \end{tabular}
\end{table}

The ordering is not the one the architectural progression would predict. Pooled across the three scenes, v1 leads at an IoU of 0.5993, v2 follows at 0.5857, and the volumetric v3 trails at 0.5706, with the pre-search v3 configuration close behind at 0.5466. v3b, trained on plain RGB instead of the synthesized RED, NIR and NDVI channels v3 reads, leads all four at 0.6668.

\subsection{Theory on why v3b Wins on Synthetic Data}
\label{sec:results_v1_discussion}

v1 differs from v3 in three ways at once: two dimensions against three, ImageNet pre-training against training from scratch, and RGB against the synthesized RED, NIR and NDVI channels v3 reads instead. A win for v1 in \autoref{tab:density_sweep} cannot be pinned on any one of those three, so it is not the comparison to build a theory on. v3b isolates the third difference on its own: it is v3's architecture, optimizer, data split and preprocessing unchanged, with only the input swapped for the RGB focal stack, and it was trained precisely to answer this question directly.

Scored the same way as v3, v3b reaches a pooled IoU of 0.6668 against v3's 0.5706 (\autoref{tab:density_sweep}), well past v1's own pooled score of 0.5993. With architecture and training held fixed, RGB beats RED, NIR and NDVI outright, and by a wider margin than v1's edge over v3 in the same table.

The likely reason is what the metric rewards rather than a general advantage of RGB input. The evaluation is a per-voxel overlap on a synthetic scene rendered by the same simulator that produced the training data, and a model that latches onto simulator-specific texture statistics scores well on it. v3b can do that best, because it reads all three RGB channels directly and can fit the exact, limited color palette that the three procedural tree species and the textured ground plane produce (\autoref{sec:simulator_setup}). v3 trades two of those channels for the synthesized NIR band and the NDVI index, which removes some of the color signal along with some of the shortcut, and \autoref{fig:simulator_comparison} shows the textured ground plane reduced that shortcut rather than eliminating it. Overfitting to the generated trees' own colors, not a real advantage of measuring in RGB, is therefore the more plausible explanation for v3b's synthetic-data lead.

That lead does not carry over to a real capture. \autoref{sec:results_real} runs v3 and v3b on three real flights and finds close to the opposite: v3b's predicted volumes collapse to almost nothing in every season, while v3 keeps a crown-like shape in two of the three. The ranking in \autoref{tab:density_sweep} is a property of the simulator's own color statistics, not a preview of which channel set holds up once the input is an actual photograph.

The v2 result is informative as well. Giving a 2D network more input planes instead of three made it worse. A 2D convolution treats its input channels as an unordered set, so the extra planes arrive as nine more channels with no depth semantics attached, which leads to the assumption that the G and B channels are quite informative for the task in the simulator domain, possibly more informative than the synthesized NIR channel. The v3b comparison above supports that reading directly: restoring G and B in place of NIR and NDVI is exactly what wins on synthetic data.

\subsection{The v4 Transfer Variant}
\label{sec:results_v4}

v4 never reached volume-wide inference and therefore does not appear in \autoref{tab:density_sweep}. Its per-voxel formulation needs one forward pass per voxel, which is 193{,}600 passes for a single plane and about ten hours on the available hardware. Training was restricted to one depth layer and ran for roughly 24 hours. Layer 200 is the only result that exists.

That single layer is enough to show the transfer did not work. The predicted values on layer 200 span a range of only 34 to 57 on the 8-bit scale, which is a probability range of 0.13 to 0.22. The output never crosses 0.5, so at the standard threshold the model predicts every voxel as void and its IoU is exactly zero. Sweeping the threshold down into the range the model actually occupies does not rescue it. The best IoU over all cutoffs is 0.4994, and it is reached at a cutoff below the minimum output value, where the model predicts every voxel as matter. That number is the positive class prior of the layer, not a measure of any recovered structure.

The most favorable non-degenerate operating point is a cutoff of 46, which gives an IoU of 0.3762 at a precision of 0.4886. The positive class covers 49.9\,\% of this layer, well above the 29.6\,\% it covers over the whole volume, because layer 200 sits in the dense middle of the canopy where most of the matter is. Measured against that prior, a precision of 0.4886 is slightly below what random guessing would produce. Accuracy stays within half a percentage point of 0.50 at every threshold. The model has no discriminative power on this layer at all.

\begin{figure}[htbp]
    \centering
    \includegraphics[width=0.45\textwidth]{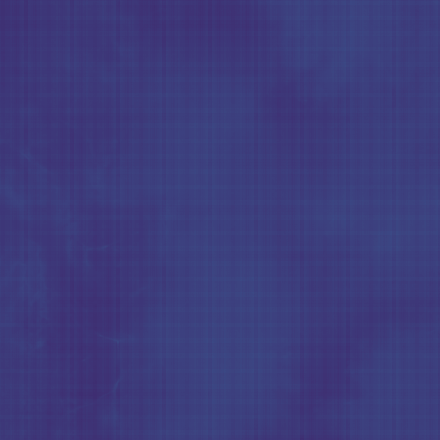}
    \hspace{0.02\textwidth}
    \includegraphics[height=0.45\textwidth]{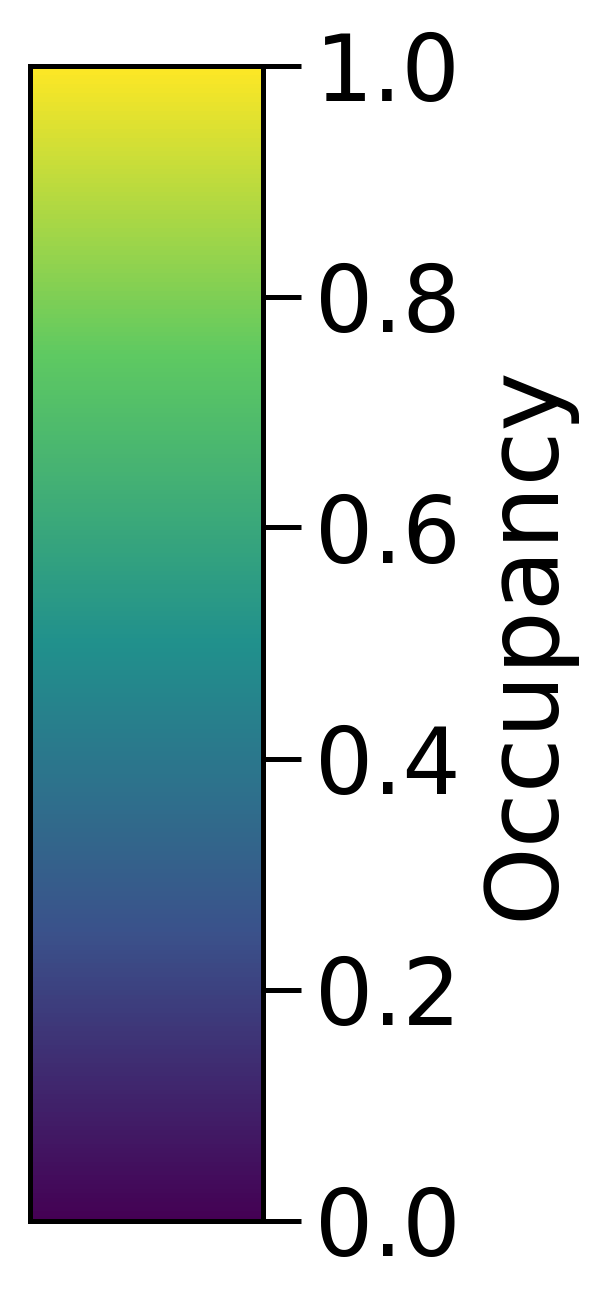}
    \caption[The single v4 output layer]{Layer 200, the only layer v4 ever produced. Shown in the viridis spectrum with no contrast stretch, so the printed appearance matches what the following stages receive. The layer spans 0.13 to 0.22 of the scale, stored as 34 to 57 of 255. The faint grid is an aliasing artifact of the PSF implementation, not structure.}
    \label{fig:v4_layer}
\end{figure}

\autoref{fig:v4_layer} is that layer. It is almost uniform. The values run from 34 to 57 on the 8-bit scale, with a mean of 46.7 and a standard deviation of 3.6, so the entire layer sits inside a band of 23 intensity levels near the bottom of the range. Nothing in it comes close to the cutoff at 128.

Two patterns are still visible. A very faint branch-like structure on the bottom left of the layer is the only thing that resembles any prediction. The other is a grid-like structure. It comes from aliasing, a known artifact of the PSF implementation.

One likely cause is the input footprint. The backbone sees a $2 \times 2$ spatial patch of per-voxel slice contributions. That is enough context to regress a reflectance value, which is what the DeepForest backbone was trained to do, and not enough to decide whether a position is inside a void. Another one is training data of the original model. Deciding that does not requires knowing what lies above and below but what lies around the voxels in question in a much narrower field. A third one could be the freezing the backbone which means the other causes cannot be corrected by training, since the only trainable part is a two-layer head on top of features that never encoded the necessary context.

\section{Effect of Hyperparameter Optimization}
\label{sec:results_hpo}

The v3 model was evaluated twice: once with the configuration used before the Optuna study, and once with the selected configuration from \autoref{tab:hpo_best}. Both runs used the same training data, the same held-out scene, and the same threshold, so the difference isolates what the search contributed.

\autoref{tab:hpo_effect} lists the two side by side.

\begin{table}[htbp]
    \centering
    \caption{The v3 architecture before and after the hyperparameter search, evaluated on \texttt{Scene\_201} at threshold 0.5. The search traded precision for recall and gained 2.4 points of IoU.}
    \label{tab:hpo_effect}
    \small
    \begin{tabular}{lrrr}
        \toprule
        \textbf{Metric} & \textbf{Before search} & \textbf{After search} & \textbf{Change} \\
        \midrule
        Precision          & 0.8052 & 0.7800 & $-0.0252$ \\
        Recall             & 0.6296 & 0.6802 & $+0.0506$ \\
        F1 / Dice          & 0.7067 & 0.7267 & $+0.0200$ \\
        Specificity        & 0.9360 & 0.9194 & $-0.0166$ \\
        Accuracy           & 0.8453 & 0.8486 & $+0.0033$ \\
        Balanced accuracy  & 0.7828 & 0.7998 & $+0.0170$ \\
        IoU                & 0.5464 & 0.5707 & $+0.0243$ \\
        \midrule
        False negatives    & 15{,}803{,}751 & 13{,}646{,}671 & $-2{,}157{,}080$ \\
        False positives    & 6{,}497{,}131  & 8{,}184{,}502  & $+1{,}687{,}371$ \\
        \bottomrule
    \end{tabular}
\end{table}

The gain is modest: 2.4 percentage points of IoU and 2.0 percentage points of Dice. The search therefore did not make the model more correct so much as it moved the operating point. The pre-search configuration was conservative. It had the highest precision and the highest specificity of any model in \autoref{tab:density_sweep}, with the downside of the worst recall. It is leaving real structure unlabeled. The selected configuration labels more voxels as occupied, catches most of what was being missed, and creates a smaller number of new mistakes.

A shortcoming about the search itself has to be recorded. Trials ran for four epochs on a 10\,\% subsample, so they ranked configurations by convergence speed rather than exclusively by converged result. A configuration that learns quickly at first and plateaus early could be favored over one that starts slower and ends better. The 2.4-point gain should therefore be read as an incremental improvement, not as the ceiling of what a full search would find.

\section{Threshold Calibration}
\label{sec:results_threshold}

The sweep described in \autoref{sec:postprocessing} was run over the full v3 prediction volume. \autoref{tab:threshold_sweep} lists the region around the maximum.

\begin{table}[htbp]
    \centering
    \caption{Threshold sweep of the v3 predictions on \texttt{Scene\_201}. Values are given on both the 8-bit storage scale and the normalized probability scale. The maximum sits at 75, well below the default cutoff of 128.}
    \label{tab:threshold_sweep}
    \small
    \begin{tabular}{rrrr}
        \toprule
        \textbf{Threshold (8-bit)} & \textbf{Probability} & \textbf{IoU} & \textbf{Dice} \\
        \midrule
        50  & 0.20 & 0.5951 & 0.7462 \\
        65  & 0.25 & 0.6062 & 0.7549 \\
        70  & 0.27 & 0.6077 & 0.7560 \\
        \textbf{75}  & \textbf{0.29} & \textbf{0.6082} & \textbf{0.7564} \\
        80  & 0.31 & 0.6079 & 0.7562 \\
        85  & 0.33 & 0.6069 & 0.7554 \\
        90  & 0.35 & 0.6052 & 0.7541 \\
        128 & 0.50 & 0.5707 & 0.7267 \\
        \bottomrule
    \end{tabular}
\end{table}

Two things stand out. The peak is worth 3.75 points of IoU over the default cutoff, which is larger than the entire gain from the hyperparameter search. And the curve is almost flat across the middle of the table: everything between thresholds 65 and 85 sits within 0.002 IoU of the maximum. It lies quite a bit below 50\,\%.
\clearpage
\section{Comparison Against Established Reconstruction Methods}
\label{sec:comparison}

In order to somehow set the performance of the 3D U-Net (v3b) into perspective the following section will try to compare it against COLMAP \cite{schoenberger2016sfm} and DepthAnything3 \cite{lin2025depthanything3}, which are two established reconstruction methods from images to 3D data. The target is a binary volume, which is the most useful representation to overlay on the output of the DeepForest framework.

Both baselines received the same input the SAI pipeline received. COLMAP ran the full pipeline on all 81 top-down captures: feature extraction with a single shared pinhole camera model, exhaustive matching, incremental mapping, and dense multi-view stereo. DepthAnything3 was run on the same 81 images in its multi-view mode, where the network predicts depth and camera pose jointly for the whole image set and the per-view predictions are fused into one point cloud. No view was discarded for either method. What differs is what each does with those views. COLMAP triangulates only the points it can match across images, so its output is limited by the correspondence search. DepthAnything3 regresses a depth value for every pixel of every view and needs no correspondences at all, so its output is dense by construction, but every one of those values still describes the first surface along a ray.

Neither method was built to return the target representation, and that has to be dealt with before any number means anything. Both return point sets, and a point set is thin where a binary volume is solid, whether it comes from Multi-View Stereo (MVS) triangulation or from fused depth predictions. Scoring such spread out points voxel-by-voxel against the dense reference is asymmetric: the baselines would be punished for the format of their output rather than for any actual reconstruction failure. A 3D dilation sweep was therefore placed in front of the comparison. It inflates the baseline outputs step by step and reports the single best score each method can reach, which removes the format penalty and tries to establish an upper bound instead of one case.

The COLMAP and DepthAnything3 outputs were pushed through the same four stages:

\subsubsection{Data ingestion}

Both baselines produce point clouds in their own coordinate frames. Each cloud was therefore first aligned by hand to the bounding box of the SAI target volume. The alignment is a mirror along the depth axis followed by a crop to a cube whose center and side lengths were set per method until the points filled the target extent. The aligned points were then mapped onto the ground truth grid of $512\times512\times550$ voxels with nearest-neighbor resampling. A voxel that contains at least one point counts as solid matter. No column filling was applied at this stage, so the voxelized clouds stay as sparse as the point sets they came from.

\subsubsection{Incremental dilation}

A full $3\times3\times3$ structuring element grows the points, following the voxel based structuring element scheme of \cite{gorte2004structuring} for tree point clouds. All 26 neighbors of a voxel are included, so this is the widest growth this kernel size allows, and it gives the sparse baselines the largest possible benefit. Computing the whole dilation again for every radius would be slow, so each step starts from the output of the last one. Memory use stays flat this way, even for 100 steps in a row.

\subsubsection{Hole filling}

Dilation on its own only grows the surface and leaves hollow shells behind. A hollow shell is a poor model of the inside of a canopy (the dense upper layer of foliage), so a 3D hole filling pass closes all enclosed inner pockets. It runs after every step, not only at the end, so each candidate volume is compared as one solid body.

\subsubsection{Global maximum search}

Each candidate was scored against the ground truth with the Jaccard Index (IoU) and the S\o{}rensen--Dice coefficient. The sweep runs from 1 to 100 steps. This draws the full accuracy curve and shows the exact step where a baseline peaks, right before the growth spills into empty space and the overlap drops again.

Both baselines are therefore measured in their own best case, tuned separately for each method. The v3b predictions get neither treatment. No dilation is applied, and no threshold calibration either: the volume is binarized at the default cutoff of $0.5$, which is the same untuned setting the architecture comparison in \autoref{tab:density_sweep} uses. That asymmetry is deliberate and it runs against v3b: \autoref{sec:postprocessing} shows that calibrating the cutoff can push a 3D U-Net's IoU well past its default-cutoff score, and no such calibration was attempted here either. Quoting the untuned value keeps the one number in the comparison that was never fitted to the scene it is reported on, and the limits of point based reconstruction stay clearly visible even so.

\begin{table}[htbp]
    \centering
    \caption[Volumetric reconstruction accuracy against the baselines]{Quantitative evaluation of volumetric reconstruction accuracy across different methodologies on Scene 201. For the point cloud baselines (COLMAP and DepthAnything3), the reported metrics reflect the absolute global maximum achieved after sweeping the optimal number of 3D morphological dilation iterations, so both are tuned on the scene they are reported on. v3b, v3's architecture retrained on the RGB focal stack (\autoref{sec:results_v1_discussion}), outputs a dense volume directly, requiring zero morphological expansion, and is entered at the default cutoff of 0.5 without any per-scene calibration. All three values come from the same metric implementation and the same nearest-neighbor resampling onto the label grid (\autoref{sec:thresholds_resampling}).}
    \label{tab:method_comparison}
    \begin{tabular}{lccc}
        \toprule
        \textbf{Method} & \textbf{Optimal Dilation Iterations} & \textbf{Peak IoU} & \textbf{Peak Dice} \\
        \midrule
        3D U-Net (v3b)   & 0  & \textbf{0.6514} & \textbf{0.7889} \\
        COLMAP (SfM)     & 40 & 0.3446 & 0.5126 \\
        DepthAnything3  & 79 & 0.2959 & 0.4567 \\
        \bottomrule
    \end{tabular}
\end{table}

The two baselines behave in different ways. DepthAnything3 gives dense surface estimates, but fusing 81 of them does not put the points on the true volume boundaries, and dilation only pushes them further into empty space. Each view contributes the canopy surface it happens to see, so the fused cloud is a thicker shell rather than a filled interior. Its best score needs 79 steps, which is already the answer: the volume only overlaps the ground truth once it has grown so far that it no longer looks like a canopy. The IoU never gets past $0.2959$.

COLMAP behaves in the following way. Only few feature points survive the matching stage, but those points are accurate and lie inside the real structures, so growing them works well. The peak comes after 40 steps at an IoU of $0.3446$, higher than DepthAnything3 and with about half the growth.

The number of steps is itself a measurement. One dilation step with the $3\times3\times3$ element grows a structure by one voxel in every direction, so 40 steps means COLMAP's points had to be grown by 40 voxels before they filled the reference. On a grid where the whole stand spans 512 voxels, that is a growth radius of about 8\,\% of the scene width around every surviving point. The gap between the recovered points and the real volume is therefore large, and what the dilation measures is mostly how far apart the points are, not how well they were placed.

Neither method recovers the inner canopy geometry, and the extra views do not change that. DepthAnything3 hallucinates a closed outer shell, and COLMAP's keypoint matching breaks down on the repeating leaf textures and the heavy self-occlusion of these scenes \cite{iglhaut2019structure}. Both are bound to what a camera can see: one depth per pixel in the first case, one match per keypoint pair in the second, and an occluded branch inside the canopy has neither. The 3D U-Net (v3b) avoids this problem instead of correcting it. It works directly on the volumetric reflectance stacks, where the 81 views have already been combined into occupancy evidence per voxel rather than into per-pixel depth by a mathematical equation. It reaches an IoU of $0.6514$ at the default cutoff, nearly double the COLMAP ceiling of $0.3446$, and COLMAP needed 40 dilation steps to get to that ceiling while v3b needed no post-processing at all.

One reservation still applies to \autoref{tab:method_comparison}: it is not a generalization estimate. Both baseline numbers were obtained by tuning the dilation step count on the same scene they are reported on, so both are upper bounds rather than expected performance on an unseen scene. The v3b entry is the only one that carries no such tuning, which is what makes the gap significant. But the scene is a single held-out scene from the same simulator that produced the training data, and the hyperparameters behind the v3b entry were tuned on validation scenes from that same simulator for v3, whose configuration v3b reuses unchanged (\autoref{sec:results_v1_discussion}). The comparison sets the v3b result in perspective rather than establishing a benchmark.

\begin{figure}[htbp]
    \centering
    \newcommand{\voxpanel}[1]{\includegraphics[trim=35 45 60 15, clip, width=\linewidth]{#1}}
    \newlength{\voxpanelht}

    \makebox[\textwidth][c]{\begin{minipage}{\dimexpr\textwidth+3cm\relax}
    \centering
    \setlength{\voxpanelht}{0.2414\linewidth}

    \begin{subfigure}[t]{0.23\linewidth}
        \centering
        \voxpanel{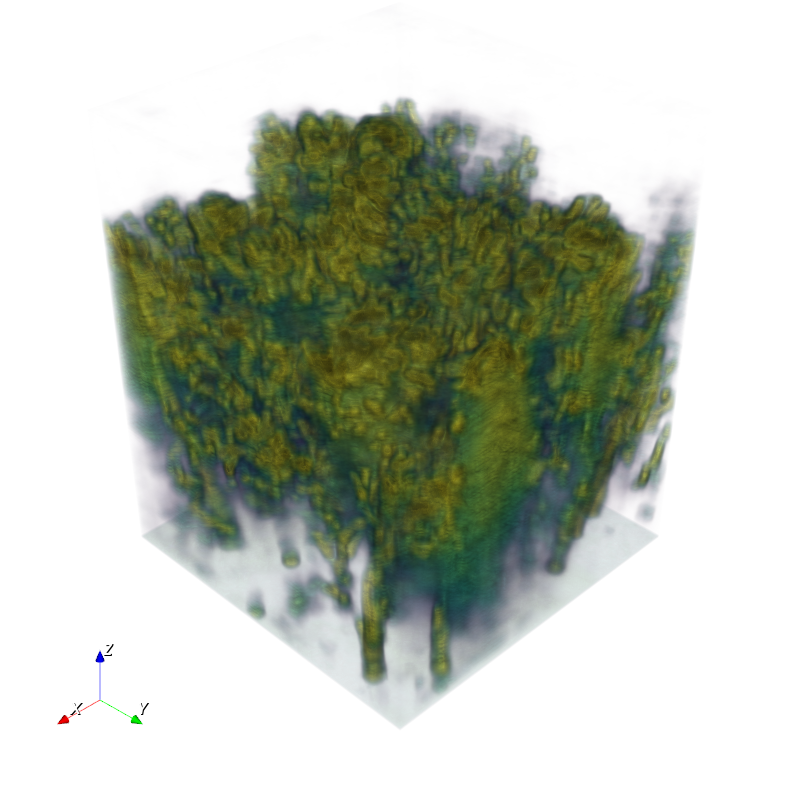}
        \caption{v3b, raw output}
        \label{fig:v3b_raw}
    \end{subfigure}\hspace{0.02\linewidth}%
    \begin{subfigure}[t]{0.23\linewidth}
        \centering
        \voxpanel{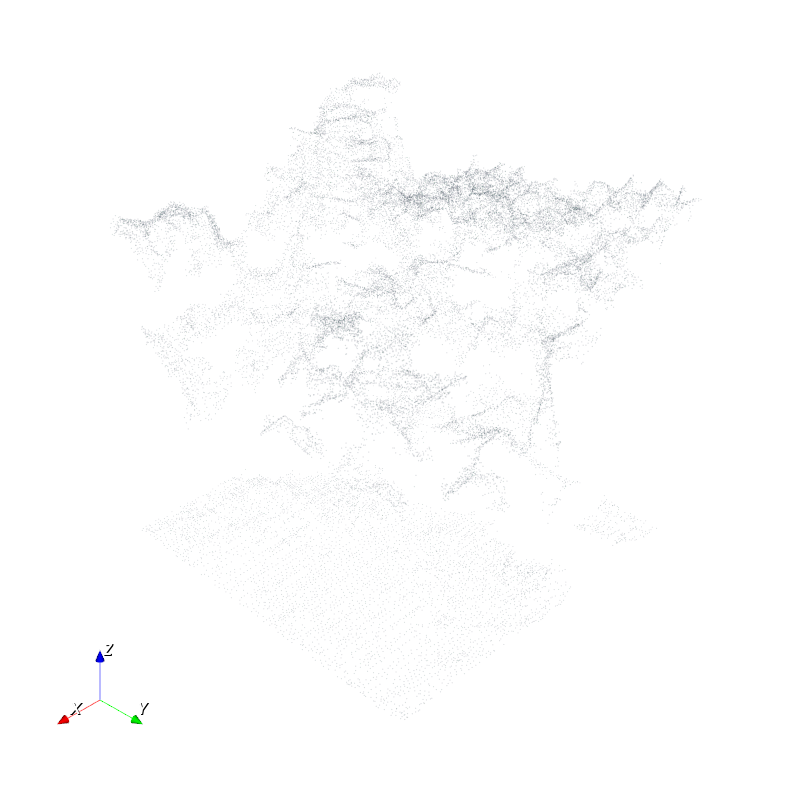}
        \caption{COLMAP, raw voxels}
        \label{fig:colmap_sparse}
    \end{subfigure}\hspace{0.02\linewidth}%
    \begin{subfigure}[t]{0.23\linewidth}
        \centering
        \voxpanel{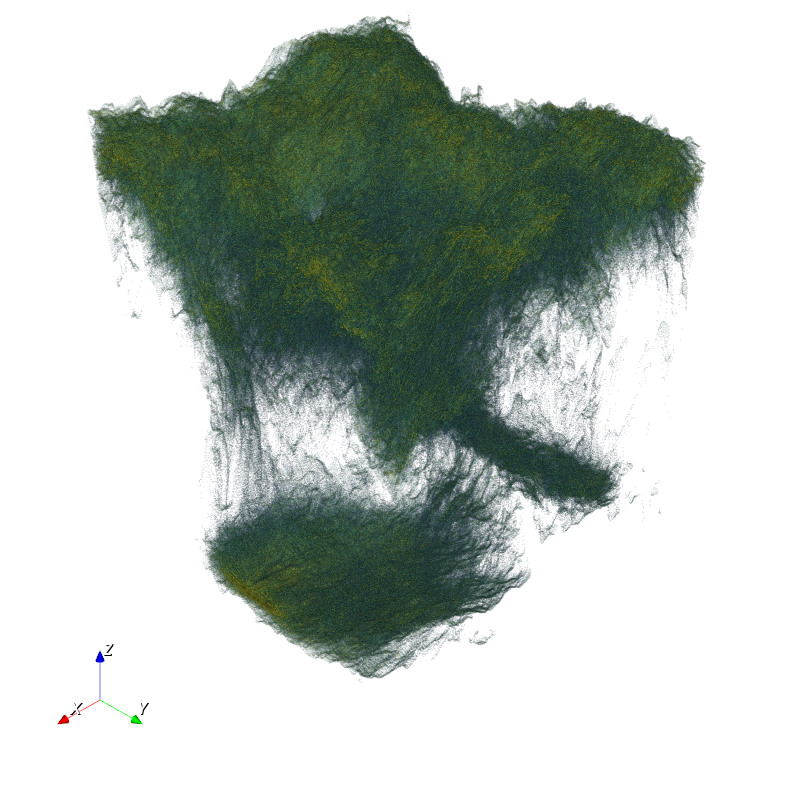}
        \caption{DepthAnything3, raw voxels}
        \label{fig:depthanything_sparse}
    \end{subfigure}\hspace{0.02\linewidth}%
    \begin{subfigure}[t]{0.23\linewidth}
        \centering
        \includegraphics[height=\voxpanelht]{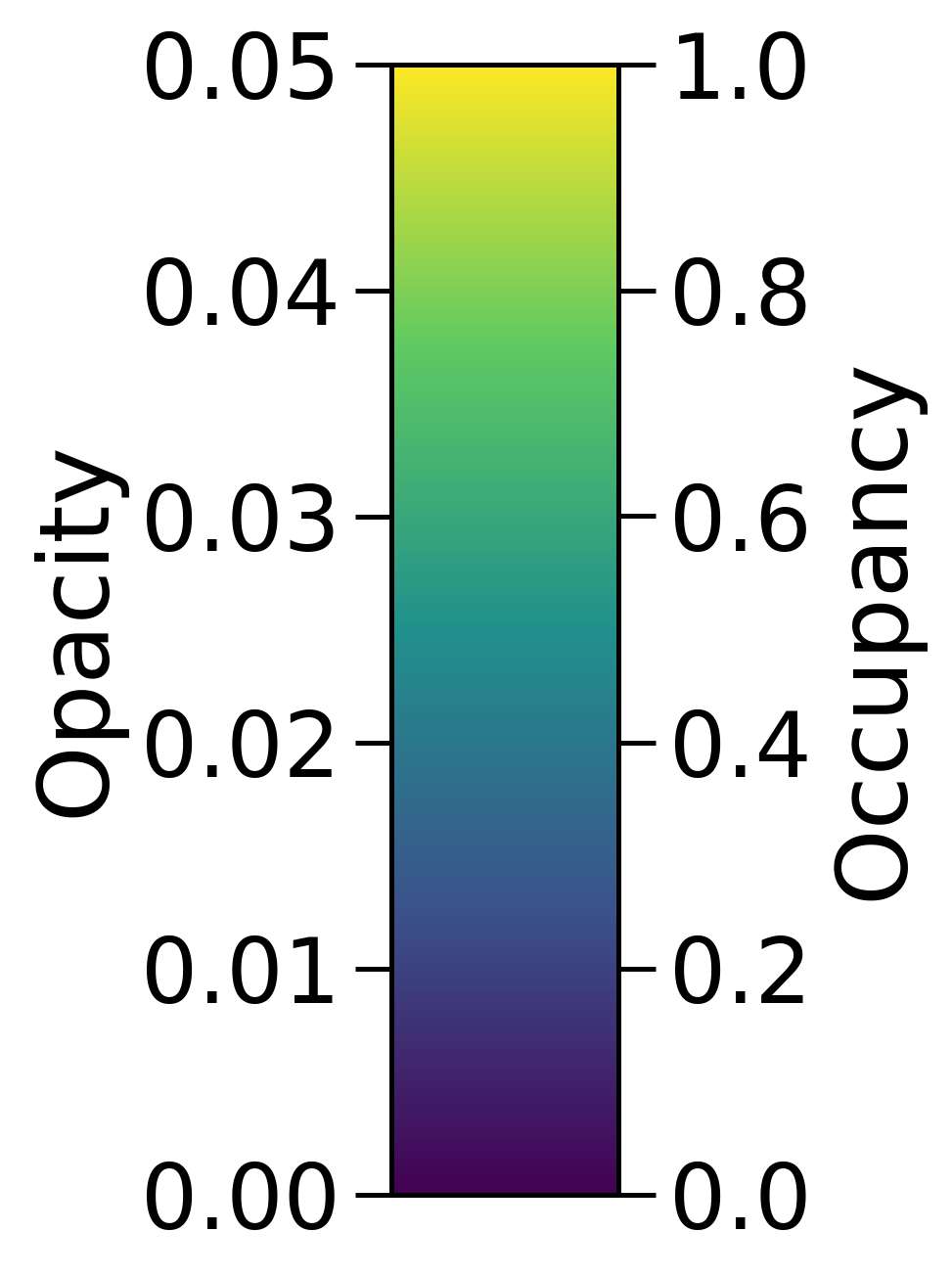}
    \end{subfigure}

    \vspace{0.8em}

    \begin{subfigure}[t]{0.23\linewidth}
        \centering
        \voxpanel{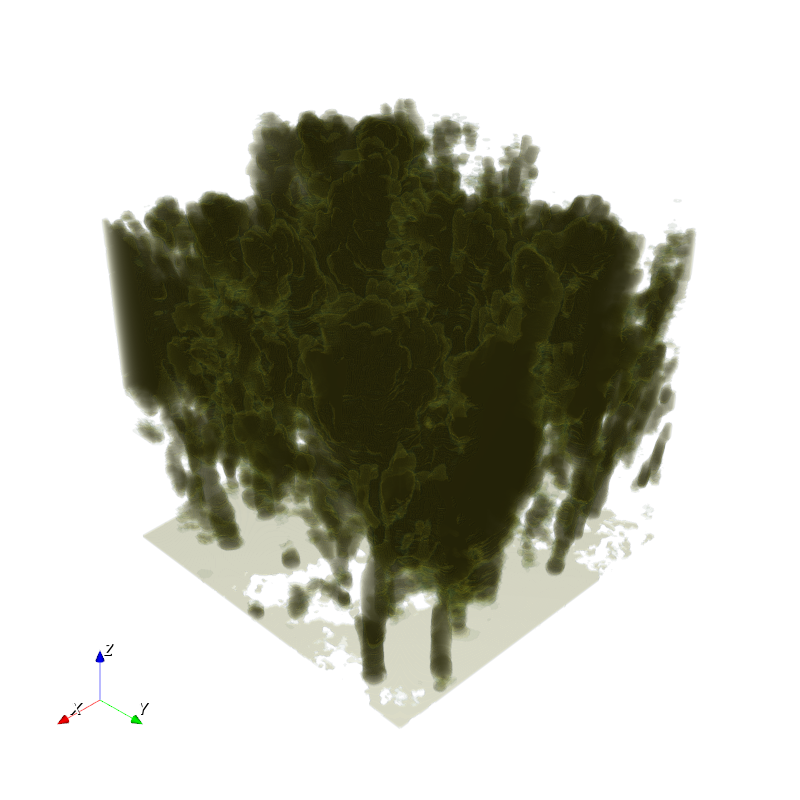}
        \caption{v3b, binarized at 0.5}
        \label{fig:v3b_binarized}
    \end{subfigure}\hspace{0.02\linewidth}%
    \begin{subfigure}[t]{0.23\linewidth}
        \centering
        \voxpanel{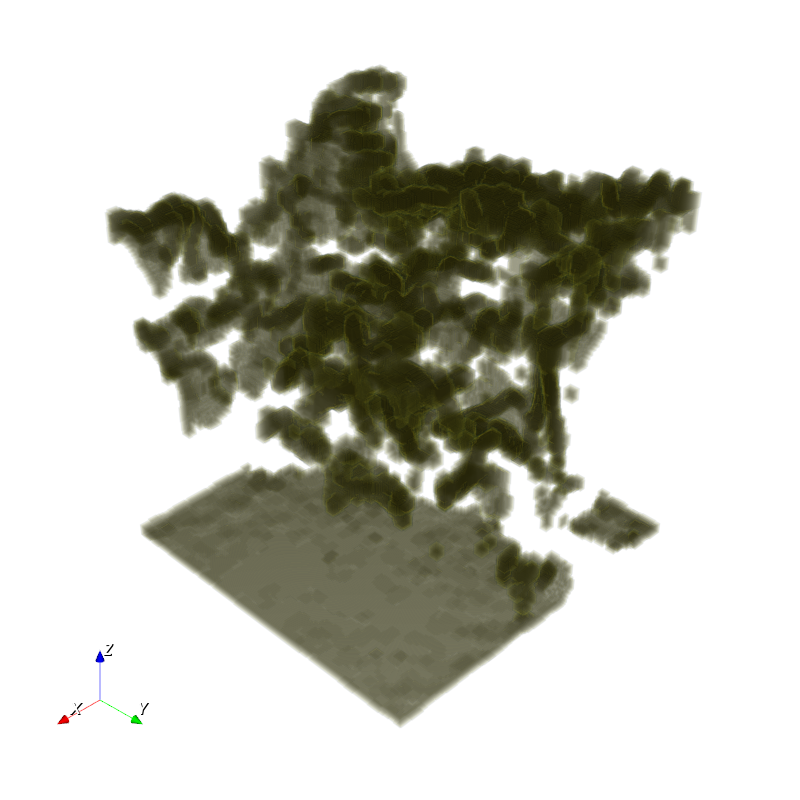}
        \caption{COLMAP, dilated}
        \label{fig:colmap_dilated}
    \end{subfigure}\hspace{0.02\linewidth}%
    \begin{subfigure}[t]{0.23\linewidth}
        \centering
        \voxpanel{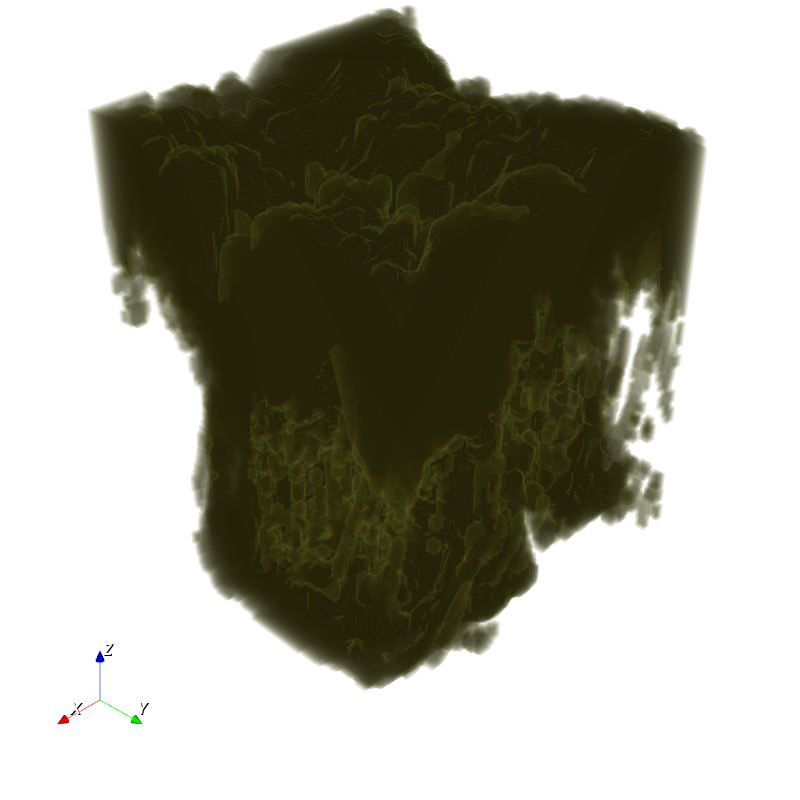}
        \caption{DepthAnything3, dilated}
        \label{fig:depthanything_dilated}
    \end{subfigure}\hspace{0.02\linewidth}%
    \begin{subfigure}[t]{0.23\linewidth}
        \centering
        \voxpanel{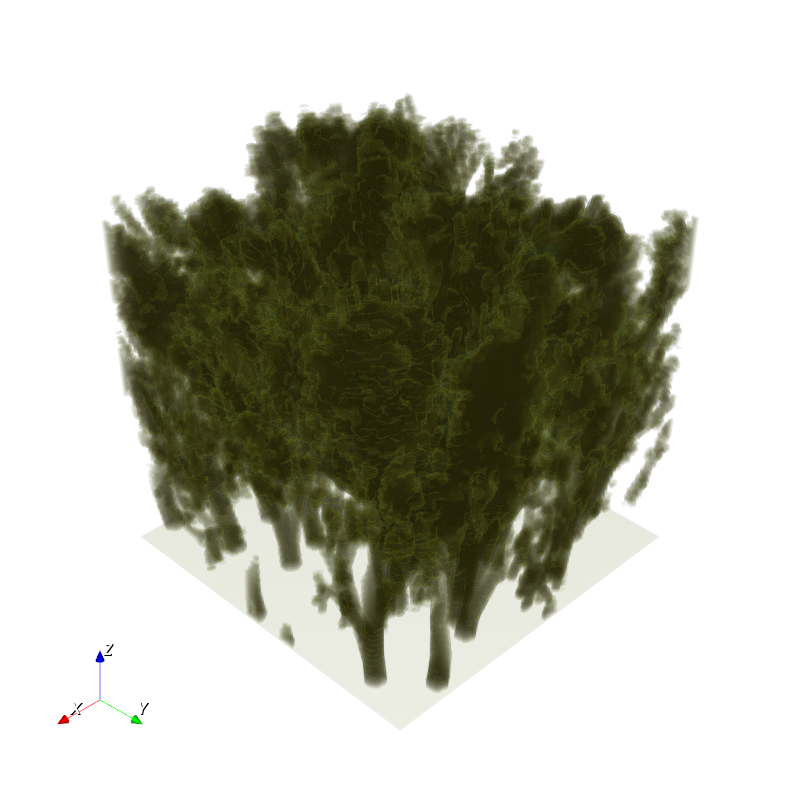}
        \caption{Processed ground truth}
        \label{fig:gt}
    \end{subfigure}
    \end{minipage}}

    \caption[Volumetric comparison of baselines, prediction and ground truth]{Volumetric structures for Scene 201. One method per column, raw output in the top row and processed volume in the bottom row. The ground truth has no raw counterpart, so the fourth slot of the top row holds the color scale. Occupancy runs from 0 for empty space to 1 for a fully occupied voxel, and the second axis gives the opacity the renderer maps that occupancy to. All panels share the same camera pose. Both baselines are shown after seven dilation steps, a middle setting chosen for readability rather than the best step count per method from \autoref{tab:method_comparison}. The v3b volume is binarized at the default 0.5 boundary, the same setting \autoref{tab:method_comparison} reports. The opacity axis holds for every panel except (b) and (c). Those two raw baseline outputs are so sparse that they are close to invisible at a peak opacity of 0.05, so they are rendered at 0.2 instead. Apparent density should therefore not be compared between those two panels and the rest.}
    \label{fig:volumetric_comparison}
\end{figure}

\autoref{fig:volumetric_comparison} puts the three methods next to each other, one method per column. The top row holds the raw output, the bottom row the processed volume. All panels use the same camera pose, so the shapes can be compared directly. The opacity setting is not the same across rows: the two raw baseline panels use a peak opacity of $0.2$ so that their sparse points stay visible at all, while every processed volume uses $0.05$. Comparing density between the top and bottom row is therefore not meaningful, comparing shape is.

The raw panels already explain the numbers in \autoref{tab:method_comparison}. COLMAP returns so few points that its volume is nearly empty, while DepthAnything3 returns a dense surface that closes the canopy from the outside. After dilation, COLMAP forms separate blobs around real branches and DepthAnything3 turns into one solid block. The v3b volume keeps single trunks and the empty space between them, before and after binarization, and that is what the higher IoU measures.

\clearpage

\section{Qualitative Behavior on Real Captures}
\label{sec:results_real}

Three real drone campaigns exist over the same forest patch: October 2024, at 48.3328\,N, 14.3313\,E, followed by a March 2025 re-flight over the leaf-off winter canopy and a May 2025 re-flight at full leaf-on. Each covers a roughly 30$\times$30\,m grid with the planned nine-by-nine camera constellation used for the simulated scenes. All three were processed into focal stacks by the same LFR pipeline used for the synthetic data, 440 planes deep.

This section reports no metrics. Quantitative evaluation needs per-voxel ground truth, and for a real forest interior that ground truth cannot be obtained. The three real campaigns contain the focal stacks and the model predictions, and no reference volume. What follows is therefore a description of observed behavior, backed where possible by numbers computed directly from the stacks, not a measured score.

\begin{figure}[htbp]
    \centering
    \newlength{\realpanelht}
    \setlength{\realpanelht}{0.34\textwidth}

    \begin{subfigure}[t]{0.34\textwidth}
        \centering
        \includegraphics[width=\linewidth]{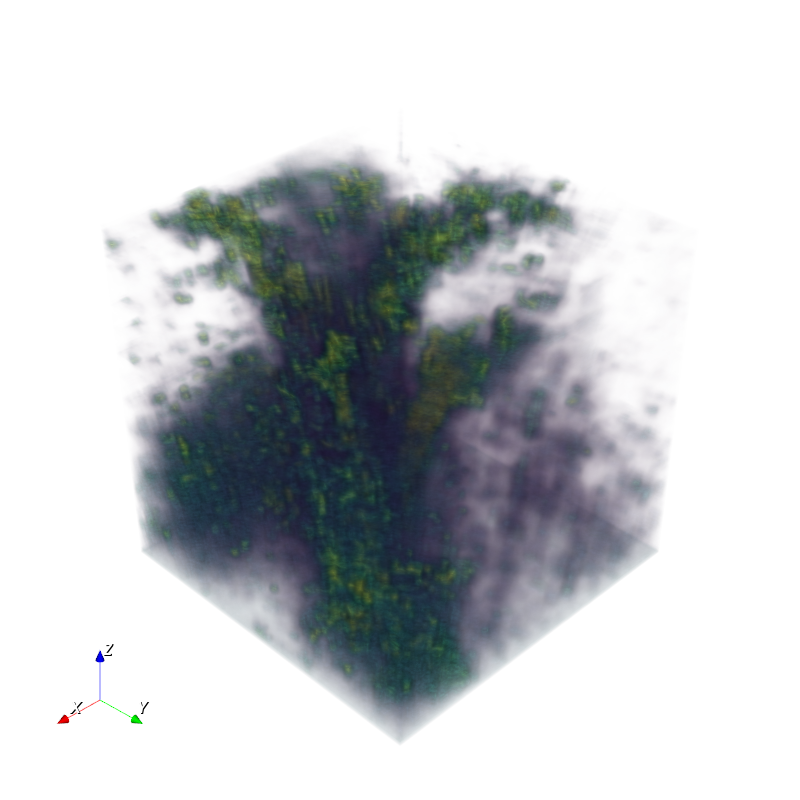}
        \caption{v3, oblique}
        \label{fig:real_v3_oblique_oct}
    \end{subfigure}\hfill
    \begin{subfigure}[t]{0.34\textwidth}
        \centering
        \includegraphics[width=\linewidth]{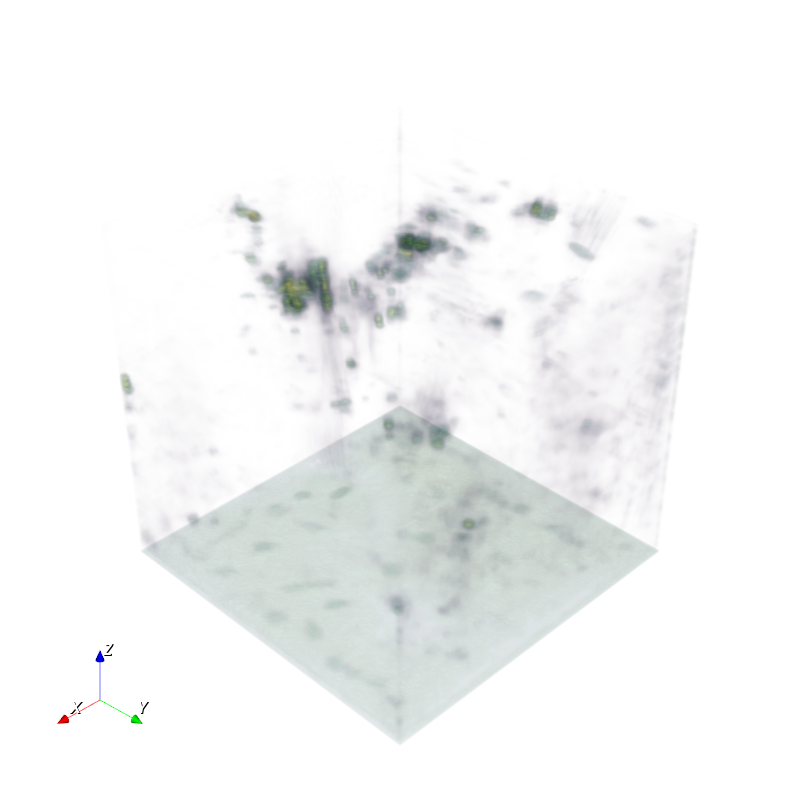}
        \caption{v3b, oblique}
        \label{fig:real_v3b_oblique_oct}
    \end{subfigure}\hfill
    \begin{subfigure}[t]{0.26\textwidth}
        \centering
        \includegraphics[height=\realpanelht]{renders/occupancy.scale.png}
    \end{subfigure}

    \vspace{0.8em}

    \begin{subfigure}[t]{0.34\textwidth}
        \centering
        \includegraphics[width=\linewidth]{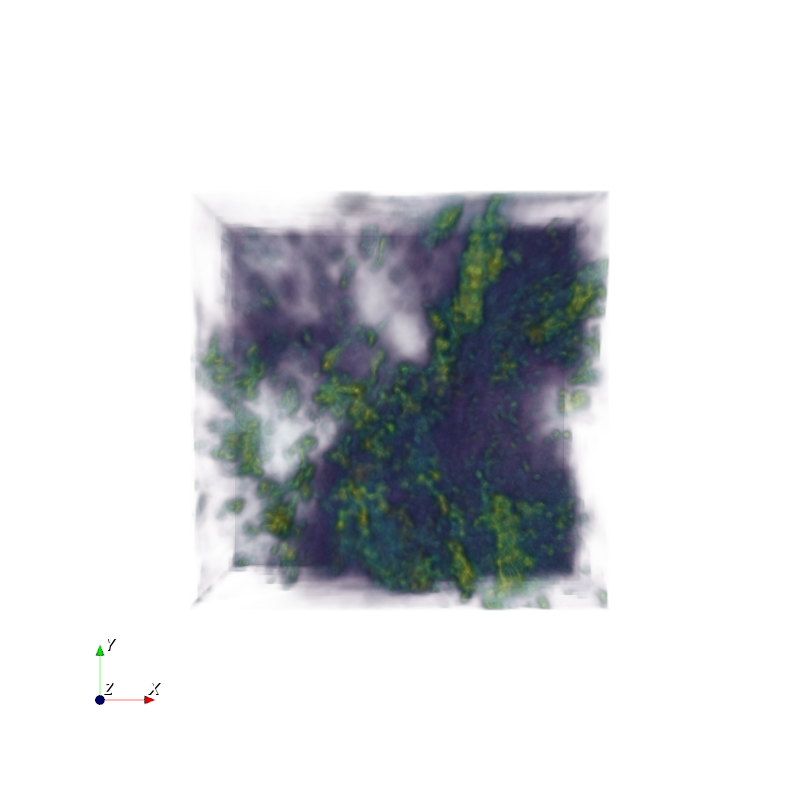}
        \caption{v3, top-down}
        \label{fig:real_v3_top_oct}
    \end{subfigure}\hfill
    \begin{subfigure}[t]{0.34\textwidth}
        \centering
        \includegraphics[width=\linewidth]{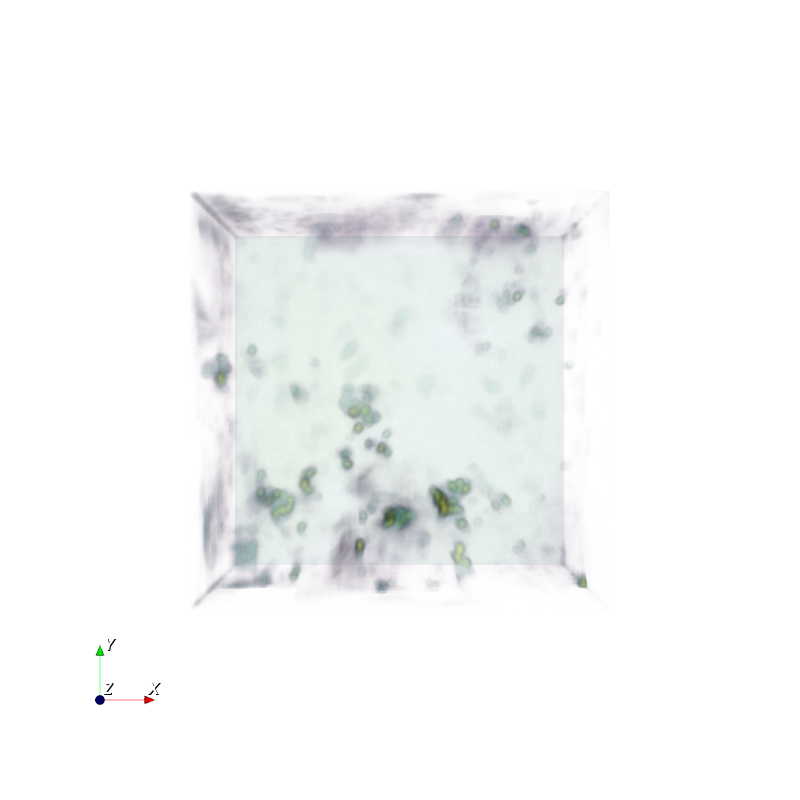}
        \caption{v3b, top-down}
        \label{fig:real_v3b_top_oct}
    \end{subfigure}\hfill
    \begin{subfigure}[t]{0.26\textwidth}
        \centering
        \reflectbox{\includegraphics[width=\linewidth,trim={3cm 3cm 3cm 3cm},clip,angle=90]{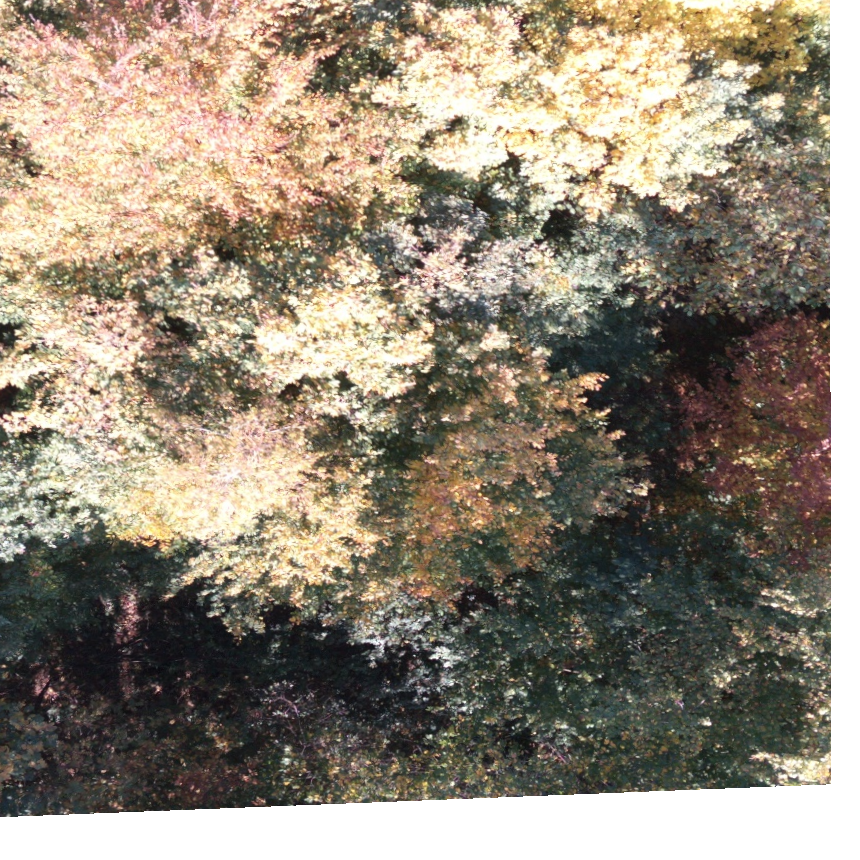}}
        \caption{center view}
        \label{fig:real_center_oct}
    \end{subfigure}

    \caption[Predicted volumes for v3 and v3b on the October real capture]{Predicted volumes for v3 and v3b on \texttt{Real\_Scene\_Oct}, $440 \times 440 \times 440$ voxels each. Left column: v3, reading the measured red and near-infrared bands. Right column: v3b, v3's architecture reading the measured RGB band instead. Top row is an oblique view, bottom row looks straight down the depth axis. Both columns come from the same October 2024 flight; the only difference is which measured channels each model reads. All panels use the same camera pose and the color scale on the right, where occupancy runs from 0 for empty space to 1 for a fully occupied voxel and the second axis gives the matching opacity, which peaks at 0.05. Bottom right, below the scale: the raw nadir photograph closest to the center of the flight grid, given for a plausibility check against the reconstructed structure above it. No ground truth exists for this capture, so no metric is reported.}
    \label{fig:real_volumes_oct}
\end{figure}

\begin{figure}[htbp]
    \centering
    \newlength{\realpanelhtmar}
    \setlength{\realpanelhtmar}{0.34\textwidth}

    \begin{subfigure}[t]{0.34\textwidth}
        \centering
        \includegraphics[width=\linewidth]{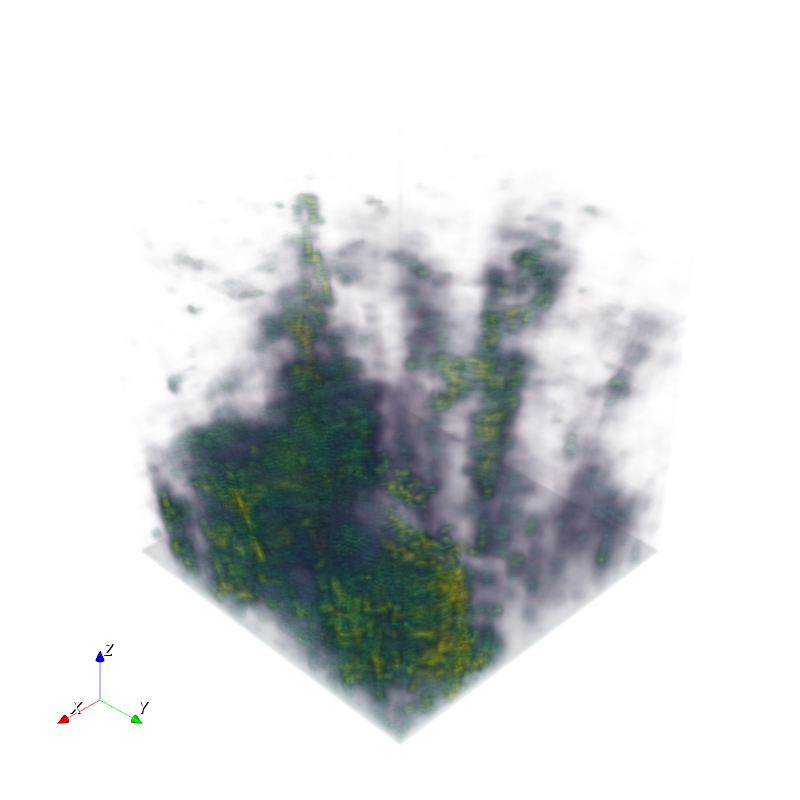}
        \caption{v3, oblique}
        \label{fig:real_v3_oblique_mar}
    \end{subfigure}\hfill
    \begin{subfigure}[t]{0.34\textwidth}
        \centering
        \includegraphics[width=\linewidth]{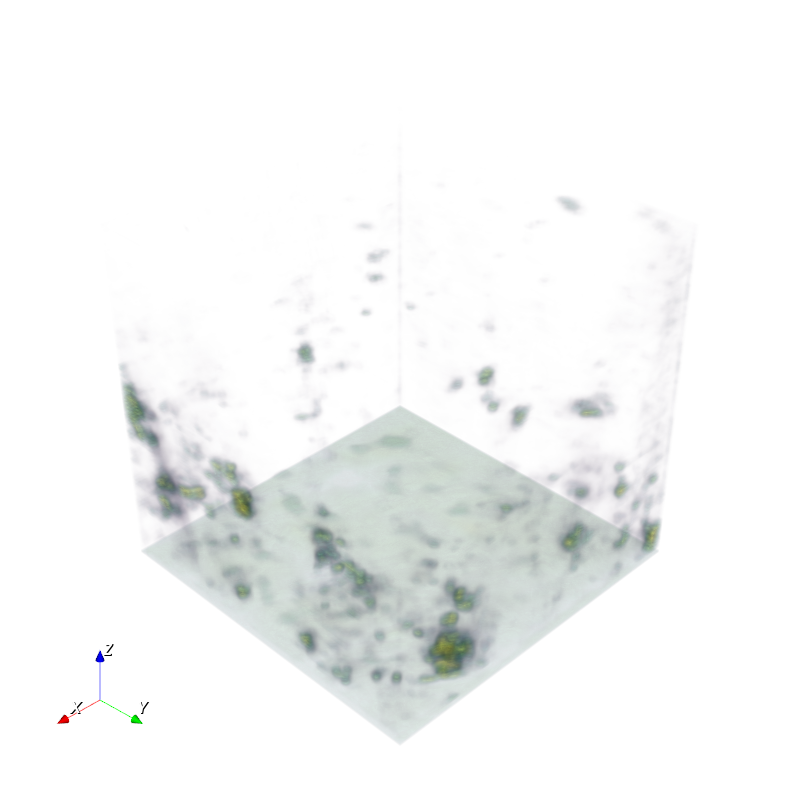}
        \caption{v3b, oblique}
        \label{fig:real_v3b_oblique_mar}
    \end{subfigure}\hfill
    \begin{subfigure}[t]{0.26\textwidth}
        \centering
        \includegraphics[height=\realpanelhtmar]{renders/occupancy.scale.png}
    \end{subfigure}

    \vspace{0.8em}

    \begin{subfigure}[t]{0.34\textwidth}
        \centering
        \includegraphics[width=\linewidth]{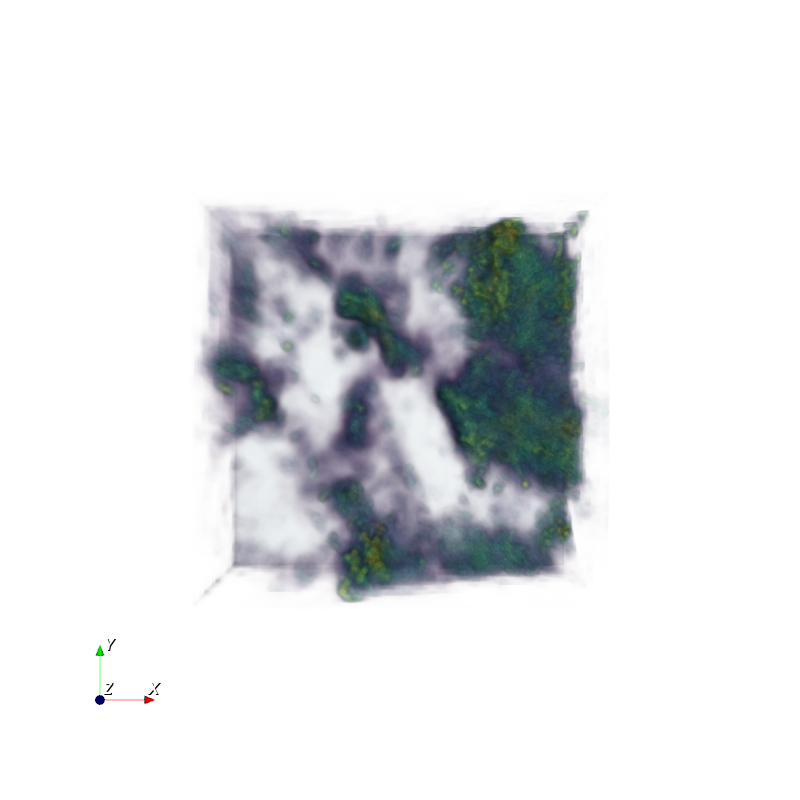}
        \caption{v3, top-down}
        \label{fig:real_v3_top_mar}
    \end{subfigure}\hfill
    \begin{subfigure}[t]{0.34\textwidth}
        \centering
        \includegraphics[width=\linewidth]{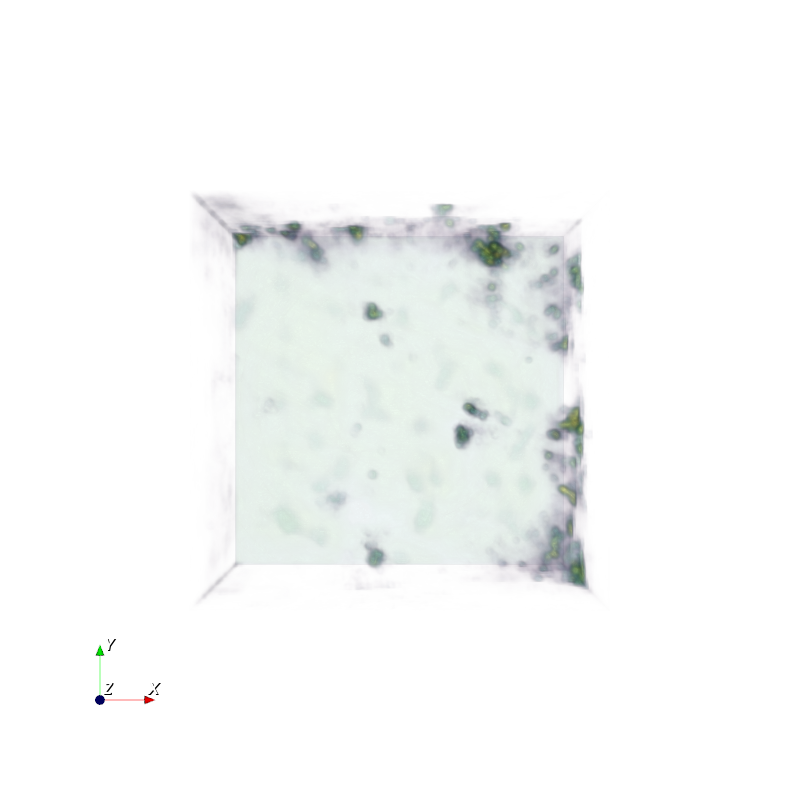}
        \caption{v3b, top-down}
        \label{fig:real_v3b_top_mar}
    \end{subfigure}\hfill
    \begin{subfigure}[t]{0.26\textwidth}
        \centering
        \includegraphics[width=\linewidth,angle=270,origin=c,trim={3cm 3cm 3cm 3cm},clip]{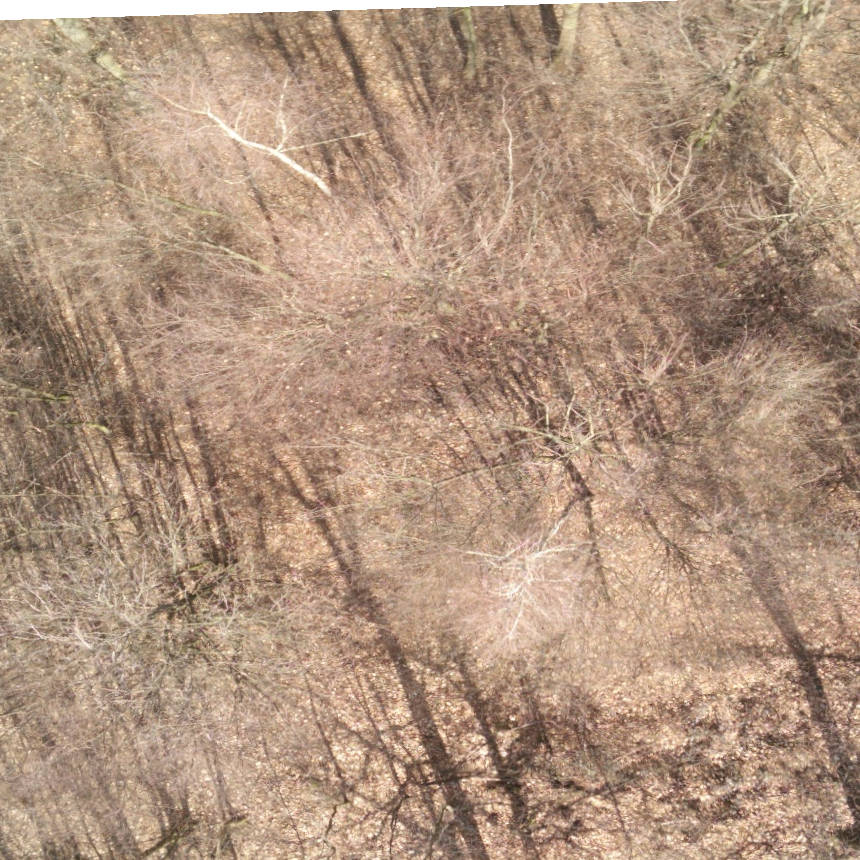}
        \caption{center view}
        \label{fig:real_center_mar}
    \end{subfigure}

    \caption[Predicted volumes for v3 and v3b on the March real capture]{Predicted volumes for v3 and v3b on \texttt{Real\_Scene\_Mar}, same layout, same camera pose and color scale as \autoref{fig:real_volumes_oct}. Both columns come from the same March 2025 flight, the leaf-off winter canopy. Bottom right, below the scale: the raw nadir photograph closest to the center of the flight grid. No ground truth exists for this capture, so no metric is reported.}
    \label{fig:real_volumes_mar}
\end{figure}

\begin{figure}[htbp]
    \centering
    \newlength{\realpanelhtmay}
    \setlength{\realpanelhtmay}{0.34\textwidth}

    \begin{subfigure}[t]{0.34\textwidth}
        \centering
        \includegraphics[width=\linewidth]{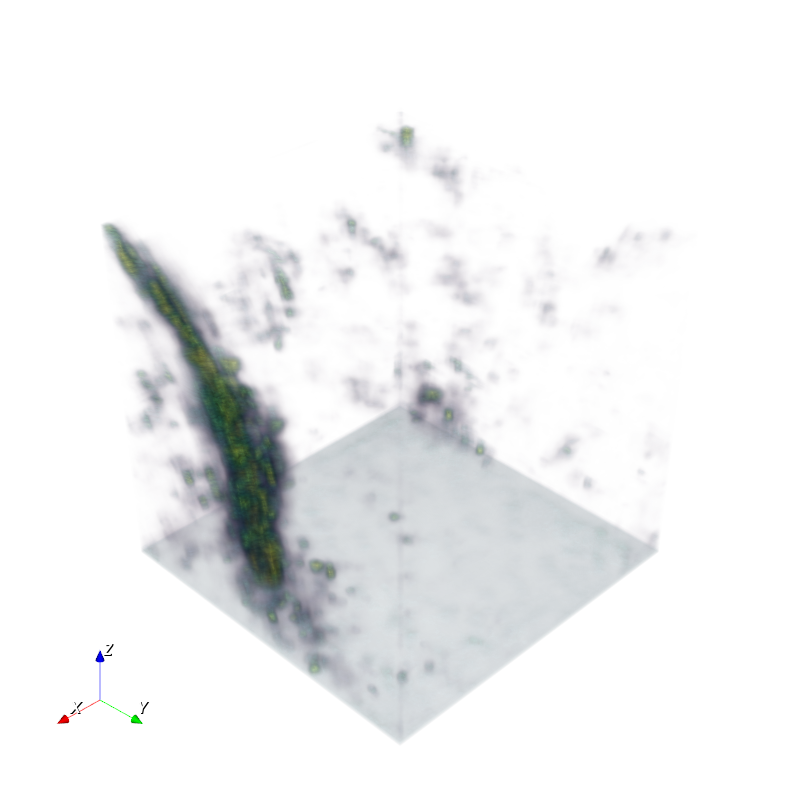}
        \caption{v3, oblique}
        \label{fig:real_v3_oblique_may}
    \end{subfigure}\hfill
    \begin{subfigure}[t]{0.34\textwidth}
        \centering
        \includegraphics[width=\linewidth]{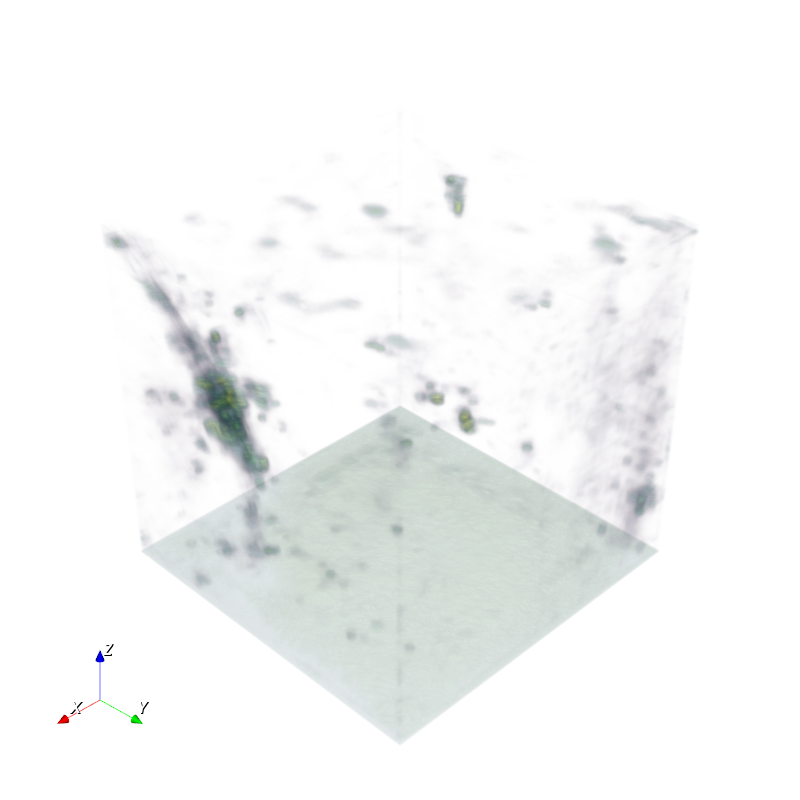}
        \caption{v3b, oblique}
        \label{fig:real_v3b_oblique_may}
    \end{subfigure}\hfill
    \begin{subfigure}[t]{0.26\textwidth}
        \centering
        \includegraphics[height=\realpanelhtmay]{renders/occupancy.scale.png}
    \end{subfigure}

    \vspace{0.8em}

    \begin{subfigure}[t]{0.34\textwidth}
        \centering
        \includegraphics[width=\linewidth]{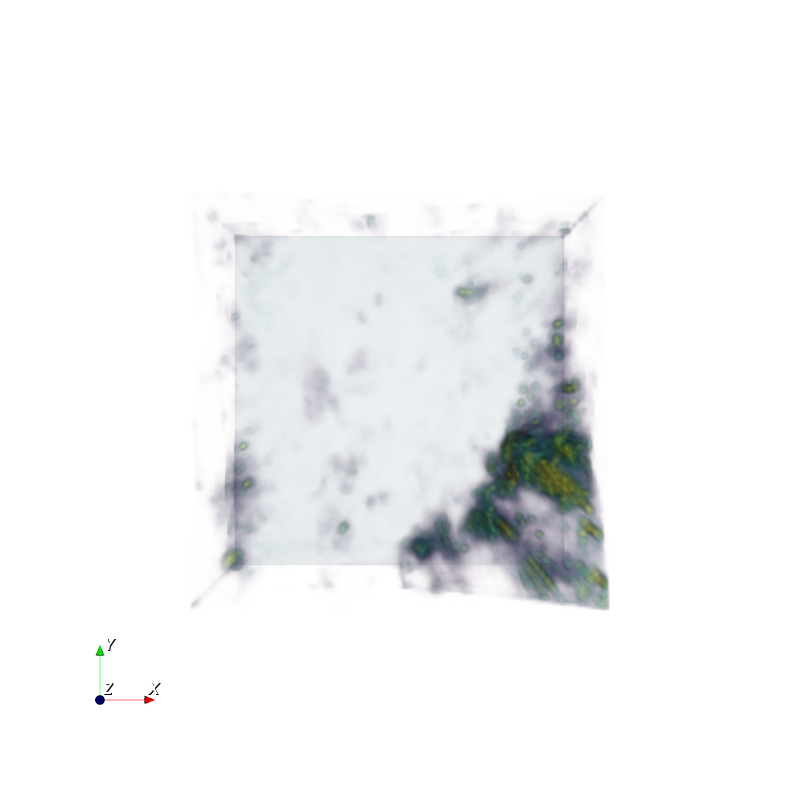}
        \caption{v3, top-down}
        \label{fig:real_v3_top_may}
    \end{subfigure}\hfill
    \begin{subfigure}[t]{0.34\textwidth}
        \centering
        \includegraphics[width=\linewidth]{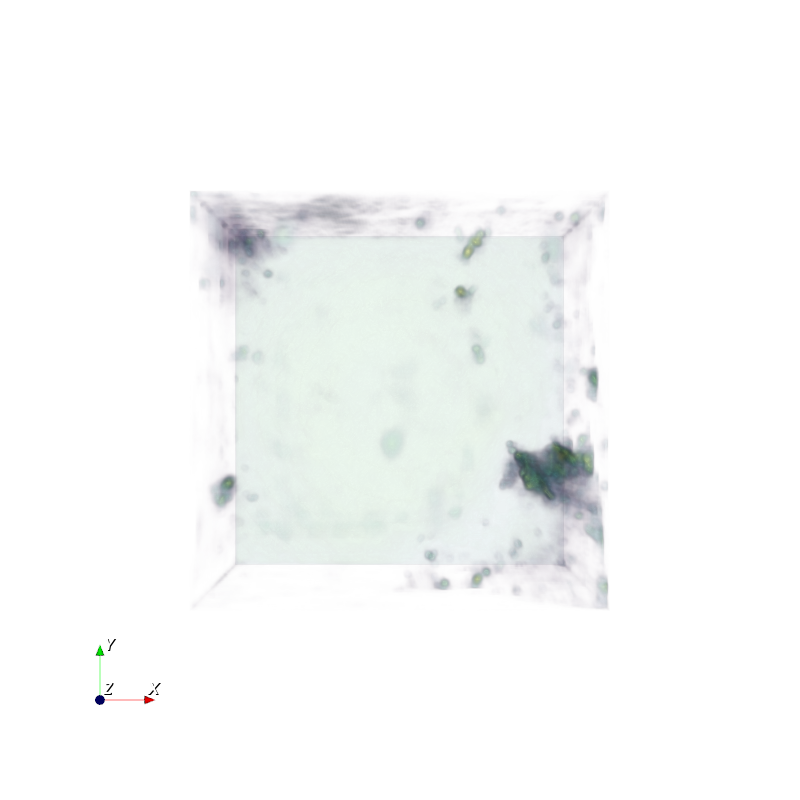}
        \caption{v3b, top-down}
        \label{fig:real_v3b_top_may}
    \end{subfigure}\hfill
    \begin{subfigure}[t]{0.26\textwidth}
        \centering
        \includegraphics[width=\linewidth,angle=0,trim={3cm 3cm 3cm 3cm},clip]{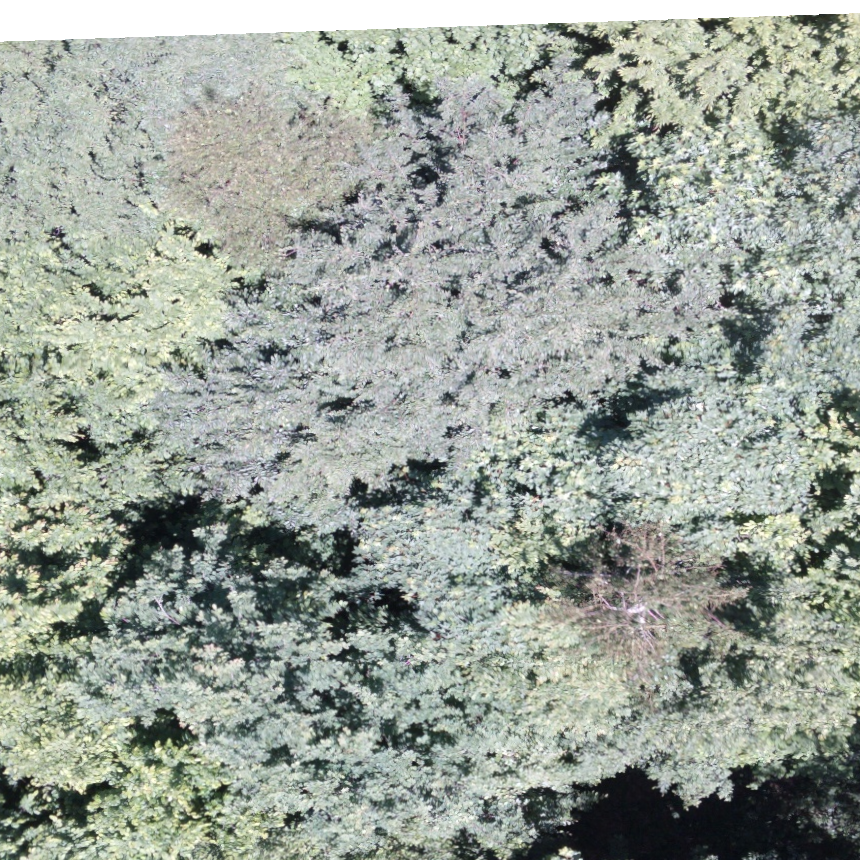}
        \caption{center view}
        \label{fig:real_center_may}
    \end{subfigure}

    \caption[Predicted volumes for v3 and v3b on the May real capture]{Predicted volumes for v3 and v3b on \texttt{Real\_Scene\_May}, same layout, same camera pose and color scale as \autoref{fig:real_volumes_oct}. Both columns come from the same May 2025 flight, the full leaf-on canopy. Bottom right, below the scale: the raw nadir photograph closest to the center of the flight grid. No ground truth exists for this capture, so no metric is reported.}
    \label{fig:real_volumes_may}
\end{figure}

\autoref{fig:real_volumes_oct}, \autoref{fig:real_volumes_mar} and \autoref{fig:real_volumes_may} compare v3 and v3b for October, March and May respectively, each once from an oblique angle and once from straight above. It is a same-flight comparison in every figure: both models see the identical focal stack for that season, v3 reading NIR and RED, v3b reading RGB.

v3 keeps the behavior described in \autoref{sec:results_v1_discussion} in October and March: the predicted occupancy varies along the z-axis, large parts of the volume stay close to zero, and what remains occupied forms a shape a viewer would call a tree crown, dense near the stem base and thinning out with height. May breaks that pattern: what survives is a single dense streak hugging one edge of the cube rather than a crown-like shape spread across the footprint, consistent with the low mean occupancy in \autoref{tab:real_seasonal}.

v3b does not reach a crown shape in any of the three seasons. The volume is almost entirely empty; what little survives sits in small disconnected clumps with no crown-like shape, in October, March or May alike. Averaged over all 440 layers, v3's mean predicted occupancy on October is 0.143 against v3b's 0.011, a factor of thirteen. That is the opposite of \autoref{sec:results_v1_discussion}'s synthetic result, where v3b's architecture-matched RGB input outright beat v3's NIR, RED and NDVI channels (pooled IoU 0.6668 against 0.5706). The most likely reason is where each channel's domain gap sits. v3b was trained only on Gazebo renders, so its RGB input at test time on a real photograph is a rendering-realism gap: real drone JPEGs do not share the texture statistics of the simulator. v3's NIR channel is GAN-synthesized during training, a domain gap of a different kind, but at inference on a real campaign it is the sensor's own measured band, so the gap that matters at training time is absent at test time. Both explanations are plausible and neither is confirmed; the only measured fact is that the geometric edge RGB holds on synthetic data does not carry over to a real capture, and here it is v3's multispectral input that keeps a usable volume, at least for two of the three seasons.

\begin{table}[htbp]
    \centering
    \caption[Predicted occupancy across the three real campaigns]{Mean predicted occupancy of v3 and v3b, for all three seasons of the same flight. Both models were run over the full 440-layer stack of each campaign.}
    \label{tab:real_seasonal}
    \small
    \begin{tabular}{lllrr}
        \toprule
        \textbf{Campaign} & \textbf{Date} & \textbf{Canopy state} & \textbf{v3 density} & \textbf{v3b density} \\
        \midrule
        October & 2024-10-25 & fall & 0.143 & 0.011 \\
        March   & 2025-03-03 & leaf-off  & 0.084 & 0.012 \\
        May     & 2025-05-14 & full leaf-on  & 0.022 & 0.010 \\
        \bottomrule
    \end{tabular}
\end{table}

\autoref{tab:real_seasonal} extends both numbers to all three seasons. From a visual check of the attached center perspective in every season, the ranking of density can be estimated: May is the densest canopy, followed by October and then March. v3b's predicted density does not move with season at all, staying flat between 0.010 and 0.012 regardless of canopy state, consistent with a model that fails to find structure in a real RGB stack independent of what that structure actually looks like. v3's predicted density does move with season, but not in the direction that the visual check would suggest. It falls from October's 0.143 through March's 0.084 down to May's 0.022. A full-leaf canopy should carry more occupied volume than a leaf-off one, not less. Why v3 predicts the least occupancy for the season with the most vegetation cannot be answered with the data available.

The result of both models is not satisfactory for an application on real data. For actual use the method has to be improved further with methods described in \autoref{sec:future_work}.

\chapter{Conclusion and Future Work}
\label{chap:conclusion}

This concluding chapter summarizes the key findings of the thesis, discusses the limitations of the current approach, and proposes directions for future research.

\section{Summary}
\label{sec:summary}

This thesis presented a method to improve remote sensing of forest canopies through Synthetic Aperture Imaging (SAI) (\autoref{chap:intro}). Heavy self-occlusions in dense vegetation create severe observational challenges. To address this, a 3D U-Net architecture was designed to predict void points directly within volumetric reflectance stacks. The network filters out void regions. Models were trained using the synthetic data pipeline detailed in \autoref{sec:data_gen}.

The central change against prior work is on the label side rather than on the imaging side. DeepForest \cite{youssef2025deepforest} estimates reflectance, and a reflectance target has to be anchored on a voxel with a measured value, so empty space is not part of what that network is trained to describe. Here empty space is the target. The labels come from the depth-clipped render of the scene (\autoref{sec:ground_truth}) and the void data is kept as a negative class. Every other design decision follows from that: the task becomes segmentation instead of reflectance regression, the loss becomes binary cross-entropy over voxels, and the evaluation uses overlap metrics instead of intensity errors.

Four architectures were trained and compared on a held-out synthetic scene (\autoref{sec:results_synth}). The results were not the ones the architectural progression predicted. The plain 2D U-Net scored highest on in-domain voxel overlap at an IoU of 0.6290, ahead of the volumetric v3 at 0.5707. The multi-channel v2 scored worst of the three at 0.5465, which shows that handing a 2D network more depth planes as extra channels does not by itself buy depth reasoning. The transfer variant v4 never reached volume-wide inference and showed no discriminative power on the one layer that was trained.

v3 and v3b, each the strongest variant on its own kind of input, were the pair carried into the real-capture comparison (\autoref{sec:results_v1_discussion}). There the synthetic ranking reverses: v3b's predicted volumes collapse to almost nothing in every season, while v3 keeps a crown-like shape in two of the three (\autoref{sec:results_real}). No reference exists on real data, so that reading rests on an observation and not on a number, and the in-domain table disagrees with it.

Two secondary findings came out of the evaluation. The decision threshold matters more than the hyperparameter search did: calibrating the cutoff gained 3.75 points of IoU against 2.4 points from the Optuna study (\autoref{sec:results_threshold}). And the optimum sits near 0.29 rather than 0.5, which traces back to the dilation applied when building the ground truth and should therefore be expected on any scene labeled the same way.

The comparison against established reconstruction methods (\autoref{sec:comparison}) gave both baselines their own best operating point through a 100-step dilation sweep and still left a clear gap. COLMAP peaked at an IoU of 0.3446 after 40 dilation steps, DepthAnything3 at 0.2959 after 79, against 0.6514 for v3b at its default cutoff and with no per-scene tuning of any kind. Both baselines had all 81 views available and both fail for structural reasons rather than tuning reasons: they reconstruct visible surfaces, by matching them across views in one case and by regressing a depth per pixel in the other, and the interior of a closed canopy presents no visible surface to either. So they simply output another data format and the translation cannot be done scientifically fair.

Using simulated sensor parameters effectively bypassed the scarcity of annotated real-world volumetric data, which was the constraint that made the supervised formulation possible at all. What it did not bypass is the question of whether a model trained that way is accurate outside the simulator. On that point this work delivers a qualitative observation and not a measurement.

\section{Limitations}
\label{sec:limitations}

\subsection{No Quantitative Real-World Validation}

The most serious limitation is that the central claim of the work is untested on real data. The model is meant to label empty regions in real SAI captures, and no per-voxel reference exists for any real capture, so its accuracy there is unknown. \autoref{sec:results_real} reports observed behavior only. Every number in \autoref{chap:results} comes from simulated scenes from the training domain rendered by the same simulator that produced the training data, which is the most favorable possible test condition.

\subsection{Sim-to-Real Gap}

A primary limitation of this work is the domain gap between synthetic training data and real-world forest environments. Domain adaptation strategies (\autoref{sec:nir_gan}) reduced visual discrepancies. However, the structural variability of physical canopies---the uppermost layer of the forest formed by tree crowns---remains difficult to simulate with absolute precision. Three procedural species with a shared shape parameter set do not cover the range of real branching patterns, and nothing in the simulator produces dead wood, undergrowth, or mixed-age stands.

Unpredictable environmental factors like wind-induced motion, varying solar illumination and imprecise measurement of camera positions further complicate the simulation process. Wind is the hardest of the three, since SAI assumes the scene is static across all 81 captures and there is no way to enforce that outdoors. The countermeasures actually applied---per-slice input normalization and batch normalization inside the network---only remove the photometric part of the gap. They do nothing about structural or motion differences.

Running v3 and v3b on all three real seasons (\autoref{sec:results_real}) put a number on how uneven that gap is across input channels. v3b, the architecture-matched RGB variant that wins by a wide margin on synthetic data, predicts an almost empty volume on every one of the three real campaigns and does not react to season at all. v3's real predictions do change with season, moving from October's 0.143 through March's 0.084 down to May's 0.022, the reverse of what a visual estimation implies about the vegetation actually present. That reversal is not explained here, but the fact that a number moves with season at all is a quantifiable fact about the NIR-GAN-based approach: v3, reading the RED, NIR and NDVI channels, is the only variant that leaves any season-dependent signal on real data, while v3b's RGB input stays flat regardless of season. A synthetic benchmark score is therefore not a reliable predictor of which channel will hold up on a real capture.

\subsection{Ground Truth Is Not the True Geometry}

The label volumes are dilated by an $11\times11$ elliptical element and hole-filled before use (\autoref{sec:ground_truth}). This was necessary to give the model a stable gradient to learn, but it means every reported score measures agreement with a thickened reference, not with the actual simulated geometry. Comparisons within this thesis stay valid because all methods face the same reference, but the absolute values carry an offset to the true geometry and the process can only be reversed approximately.

\subsection{Computational Cost}

The cost of a 3D CNN is a limit on training, not on deployment. A 3D U-Net keeps every intermediate activation for the backward pass, and at $440\times440$ with four depth planes the activation memory grows quickly with batch size and feature width. The GPU used throughout this work is an NVIDIA GeForce GTX 1070 Ti with 8\,GiB of VRAM, in a machine with an AMD Ryzen 5 2600 and 16\,GiB of system memory. Early exploratory runs exhausted that 8\,GiB, which is why the hyperparameter search space was bounded at batch size 4 and 32 initial features (\autoref{sec:hpo}). Within those bounds all 20 trials completed, so the memory limit shaped the search space instead of interrupting it. What it restricts is therefore how large a configuration could be searched at all, and a wider or deeper network than the ones tried here would need different hardware.

The search itself took 2 days, 17 hours and 52 minutes of GPU time for 20 trials, with the shortest trial at 1 hour 36 minutes and the longest at 13 hours 57 minutes. That is for four epochs on a 10\,\% subsample per trial. Retraining the selected configuration on the full training set came on top of it, but that run was not timed and no wall clock for it is reported here.

Inference is a different matter. v3 predicts one plane per forward pass. A single pass over a $3 \times 4 \times 440 \times 440$ window takes 66\,ms and peaks at 323\,MiB of VRAM, and the full 550-plane volume of \texttt{Scene\_201} is assembled in 88 seconds on the same machine. Only 36 of those seconds are spent inside the network, 47 go into decoding and resizing the PNG files each window needs, and 6 into writing the output maps. Against the per-voxel formulation of v4, which needs 193{,}600 forward passes for one plane and about ten hours to finish it, that is many orders of magnitude faster, so running the trained model is not the bottleneck.

\subsection{Narrow Sensor Configuration}

The trained networks were tuned to one set of simulated sensor parameters: a fixed altitude of 35 units, a $9\times9$ grid over a 24-unit baseline, a $35^{\circ}$ field of view, and a fixed set of 550 focal planes over a fixed depth range. Generalizability across varying flight altitudes, grid spacings, or focal plane counts remains untested. Since the depth spacing of the focal stack is a direct function of these settings, and since the network's anisotropic pooling is designed around that spacing, a different flight configuration would likely require retraining rather than just re-running inference.

\subsection{Synthesized Rather Than Measured Near-Infrared}

The near-infrared channel used during training is generated by NIR-GAN from rendered RGB. It is a prediction and cannot contain information absent from its input (\autoref{sec:rw_spectral}). The verification in \autoref{sec:nir_verification} runs the generator on a real RGB capture and compares the result against the measured NIR frame of the same site. It says the translation works on real input. It says nothing about rendered input, which is the only kind the training data ever uses, and no reference NIR exists for a Gazebo scene to check that against. At inference time on real data the band is measured instead, which is the intended direction, but it means the training and inference distributions for that channel differ in a way this work did not quantify. That the void prediction does not rest on the synthesized band was checked directly, not just argued: v3b, v3's architecture retrained on the RGB focal stack with everything else unchanged, reaches a pooled IoU of 0.6668 on three held-out scenes, against 0.5706 for v3 on the same scenes (\autoref{sec:results_v1_discussion}). NIR-GAN is trained on real Sentinel-2 imagery, so the RED, NIR and NDVI channels it hands to v3 already carry some of the statistics of a real sensor, while v3b's RGB input is a pure Gazebo render with a limited, procedural color palette. v3b can therefore fit the exact colors the simulator produces, while v3 is forced to learn the underlying reflectance distribution instead of a rendering signature it cannot lean on. Running both variants on the three real campaigns (\autoref{sec:results_real}) is the first check on whether that difference actually pays off outside the simulator, and there v3 is the variant that still produces a plausible volume while v3b collapses to almost nothing.

\section{Future Work}
\label{sec:future_work}

The limitations in \autoref{sec:limitations} determine this list.

\subsection{Domain Adaptation}

Addressing the sim-to-real gap is the most direct next step. Unsupervised domain adaptation via cross-domain distribution alignment \cite{sun2024cross} could improve robustness on empirical data. This avoids the need for extensive real-world annotations, which is the binding constraint here. It operates on feature distributions rather than on pixels, so it attacks the structural part of the gap that the photometric normalization used in this work leaves untouched.

A cheaper intermediate step is stronger domain randomization inside the simulator: varying illumination direction and intensity per scene, adding undergrowth and dead wood as separate object classes, and perturbing the tree shape parameters per instance instead of sharing one parameter set across all trees. None of this needs new machinery, only more simulation time.

\subsection{Architectural Extensions}

Attention gates on the skip connections are the most direct architectural extension. The v3 network has none (\autoref{sec:v3_unet}). A gate would learn a per-voxel weight for each skip connection, letting the decoder suppress the parts of the encoder feature map that carry no information for the current voxel. Since most of a canopy volume is uninformative, this kind of gating should help exactly where it is needed most, and it is a small change to an existing architecture.

The output stage is also worth revisiting. v3 collapses its output volume with a max projection along depth, which makes it a detector and biases it toward recall on structures spanning several planes (\autoref{sec:results_v1_discussion}). Predicting the full depth window directly, and supervising it against the corresponding window of the label volume, would remove that bias and could also cut inference cost, since one forward pass would produce several output planes instead of one.

\subsection{Sensor Fusion and Broader Evaluation}

Multi-modal sensor fusion offers another technical extension. Volumetric stacks derived from SAI could be combined with active sensing modalities like LiDAR. The two are complementary in a useful way: LiDAR gives sparse but metrically exact returns, while the SAI stack gives a dense but ambiguous volume. Using LiDAR data as anchors and letting the network fill between them would attack the calibration problem from \autoref{sec:results_threshold} directly, since the anchors would fix the absolute scale that the network currently has to infer.

Integrating additional spectral bands possibly even ones with high transparency would provide complementary signals, improving the network's ability to differentiate between solid biomass and empty void spaces in highly occluded regions. Adding a red-edge band would be the obvious first choice. Red-edge indices are accurate estimators of canopy chlorophyll and nitrogen content \cite{clevers2013remote}, and both of those say something about how healthy the plant is. NDVI stops reacting once the foliage is dense enough \cite{huang2021ndvi}. In that range the red-edge bands still carry information.

The evaluation framework should also be expanded across diverse forest biomes---distinct biological communities with unique structural characteristics formed by specific physical climates. Testing against different tree species and densities is necessary to confirm the methodology's general applicability.


\printbibliography[title = {References}, heading = bibintoc]

\begin{icgAppendix}
\end{icgAppendix}

\end{document}